\PassOptionsToPackage{table}{xcolor}
\documentclass[11pt,letterpaper,logo]{mystyle}
\usepackage[numbers]{natbib}

\usepackage{amsmath,amsfonts,bm}

\def\eqref#1{equation~\ref{#1}}

\def\1{\bm{1}}

\DeclareMathAlphabet{\mathsfit}{\encodingdefault}{\sfdefault}{m}{sl}
\SetMathAlphabet{\mathsfit}{bold}{\encodingdefault}{\sfdefault}{bx}{n}

\usepackage{hyperref}
\usepackage{url}
\definecolor{darkblue}{RGB}{44,52,204}
\hypersetup{colorlinks=true,citecolor=darkblue,linkcolor=darkblue,urlcolor=darkblue}
\usepackage{booktabs}
\usepackage{multirow}
\usepackage{array}
\usepackage{adjustbox}
\usepackage{graphicx}
\usepackage[table]{xcolor}
\usepackage{amsmath}
\usepackage{amssymb}
\usepackage{CJKutf8}
\usepackage{enumitem}

\definecolor{dividergray}{RGB}{240,240,240}
\definecolor{ourslavender}{RGB}{239,237,255}

\definecolor{okgreen}{RGB}{0,140,60}
\definecolor{nored}{RGB}{200,30,30}
\newcommand{\yes}{\textcolor{okgreen}{\ding{51}}}
\newcommand{\no}{\textcolor{nored}{\ding{55}}}
\newcommand{\highlight}[1]{\noindent\textbf{#1}}
\usepackage{fontawesome5}
\usepackage{tcolorbox}
\tcbuselibrary{skins,breakable}
\colorlet{myblue}{EvolventAccent}
\definecolor{failrose}{HTML}{be123c}
\definecolor{succteal}{HTML}{047857}
\definecolor{warnamber}{HTML}{b45309}

\newtcolorbox[auto counter, number within=section]{promptbox}[2][]{%
  colback=white,
  colframe=myblue,
  width=\textwidth,
  arc=2mm,
  title={\normalsize\faInfoCircle\hspace{0.5em}#2},
  breakable,
  fonttitle=\bfseries\normalsize,
  fontupper=\small,
  drop shadow southeast,
  top=2.5mm,
  bottom=2.5mm,
  left=3.5mm,
  right=3.5mm,
  before skip=3mm,
  after skip=3mm,
  boxrule=0.5mm,
  #1
}

\usepackage{pifont}

\title{\textsc{FrontierRefactor}: Benchmarking Multi-Scale Performance-Oriented Codebase Refactoring}

\author[1*]{Shijian~Wang}
\author[1,2*$\dagger$]{Fanqing~Meng}
\author[1*]{Jiajun~Chen}
\author[1*]{Qiguang~Chen}
\author[3]{Linxin~Song}
\author[1]{Runhao~Fu}
\author[1]{Shuhang~Zheng}
\author[4]{Shuai~Shao}
\author[2]{Qizhe~Xie}
\author[5$\ddagger$]{Jieyu~Zhao}
\author[1$\ddagger$]{Mengkang~Hu}

\affil[1]{Evolvent AI}
\affil[2]{National University of Singapore}
\affil[3]{University of Southern California}
\affil[4]{Shanghai Jiaotong University}
\affil[5]{Hong Kong University of Science and Technology}
\makeatletter\def\AB@affilsep{\protect\\\protect\Affilfont}\makeatother
\affil[*]{Equal contribution}
\affil[$\dagger$]{Project leader}
\affil[$\ddagger$]{Corresponding authors}
\paperlinks{\faGithub\enspace\textbf{Code:}\enspace\href{https://github.com/evolvent-ai/FrontierRefactor}{\nolinkurl{https://github.com/evolvent-ai/FrontierRefactor}}}
\runningtitle{\textsc{FrontierRefactor}}

\begin{document}

\begin{abstract}
Autonomous codebase refactoring requires coding agents to restructure working software while strictly preserving externally observable behavior, a process driven in production by execution performance and compute efficiency rather than stylistic preference. However, existing benchmarks evaluate these requirements in isolation: refactoring benchmarks evaluate functional correctness alone while overlooking severe runtime regressions, whereas code performance benchmarks target isolated micro-kernels that ignore codebase-scale complexity. We observe that software refactoring scale is governed by the scope of the \emph{observable contract}, the caller-visible interface that fixes behavioral invariants while freeing private implementations for optimization. Building on this insight, we introduce \textsc{FrontierRefactor}, a benchmark of 60 qualified engineering tasks spanning four scales from computational kernels to full repositories across in-place optimization and cross-language rewriting. Evaluating frontier coding agents through a reference-free cascaded protocol across 1,080 execution runs reveals that correctness and speedup are largely decoupled: 41.2\% of candidate submissions satisfy functional equivalence yet fail to deliver verified performance improvements, with 19.3\% suffering active performance regressions and 21.9\% trapped in measurement noise. Furthermore, refactoring paradigms reveal orthogonal engineering bottlenecks: while cross-language rewrites secure substantial speedup dividends from native compilation upon satisfying functional checks, agents struggle with contract fidelity, dropping from a 65.0\% pass rate on modules to 32.6\% at package scale; conversely, in-place optimization reliably preserves behavioral contracts at 86.5\% but rarely achieves acceleration without physical performance profiling. Trajectory analysis demonstrates that agents frequently fall victim to uncalibrated micro-benchmarking illusions, an operational blind spot that multi-turn execution timing feedback directly resolves, raising strict success from 16.0\% to 40.0\%.
\end{abstract}

\maketitle
\thispagestyle{firststyle}

\section{Introduction}

Autonomous codebase refactoring has emerged as a crucial task for evaluating whether AI agents can restructure complex software while maintaining contractual invariants~\citep{xu2026swerefactor,promax2026,swerefactorbench2026,yan2026tensorbench}. Recent efforts have made foundational progress in evaluating modular reorganization and dependency modernizations. Yet, they overlook the single most pervasive driver of software rewriting in practice: \emph{performance-oriented refactoring}. Production engineering rarely tolerates the risk of modifying working code solely for stylistic elegance. Instead, refactoring is propelled by physical operational constraints: meeting strict latency requirements, curtailing millions of dollars in compute infrastructure, and breaking architectural bottlenecks~\citep{gpt51codexmax}.

\begin{figure*}[t]
\centering
\includegraphics[width=\textwidth]{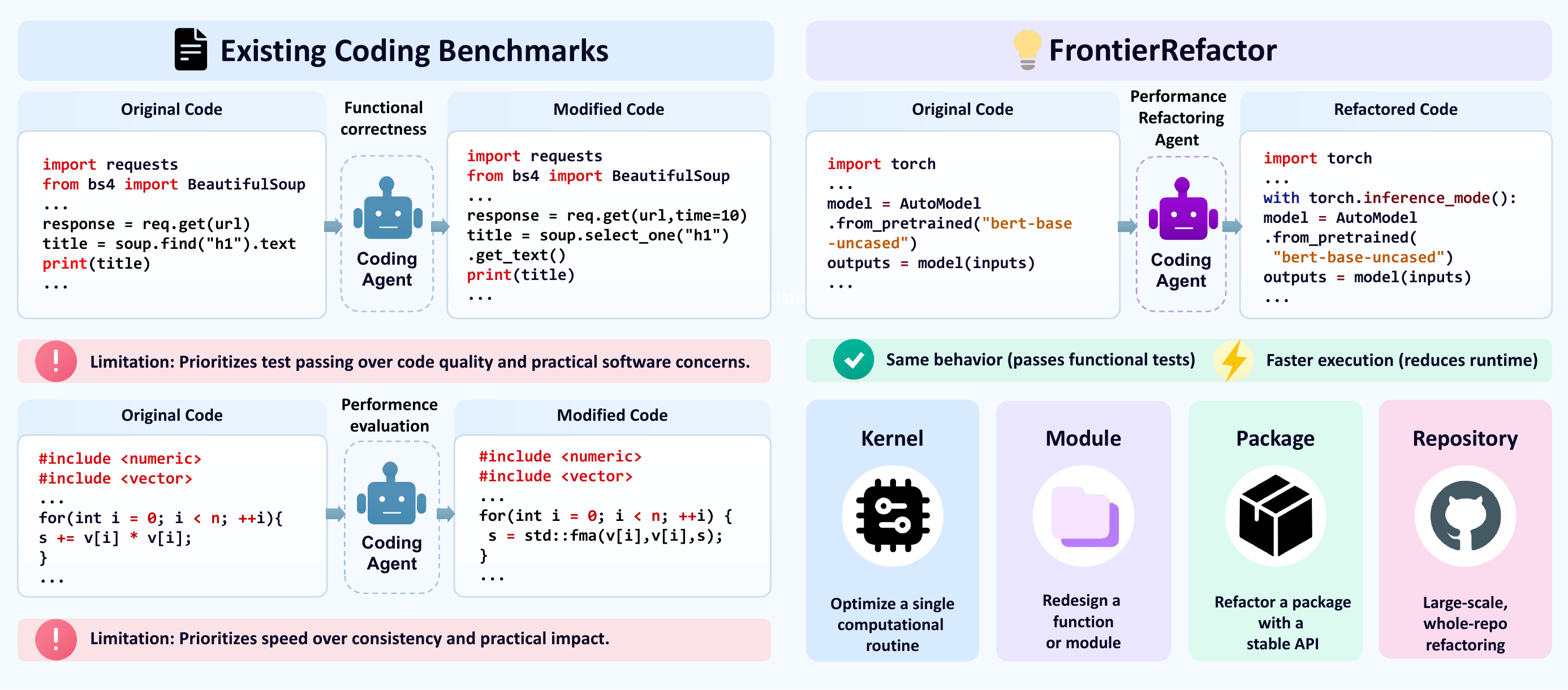}
\caption{\textbf{Conceptual comparison between existing benchmarks and \textsc{FrontierRefactor}.} \textbf{Left}: Existing coding benchmarks evaluate functional correctness alone (blind to latency regressions) or isolate narrow micro-kernels (ignoring multi-file complexity). \textbf{Right}: \textsc{FrontierRefactor} formalizes refactoring through the \emph{observable contract}, jointly enforcing behavioral equivalence and noise-calibrated speedup from kernels to repositories.}
\label{fig:teaser}
\label{fig:scales}
\end{figure*}

Evaluating coding agents on performance refactoring presents three foundational challenges (Figure~\ref{fig:teaser}):
(1)~\emph{Speed-blind refactoring vs.\ single-scale optimization}: Existing refactoring suites evaluate correctness alone~\citep{xu2026swerefactor,promax2026,swerefactorbench2026}, ignoring runtime regressions, while performance suites isolate micro-kernels~\citep{ouyang2025kernelbench,shypula2024pie,shetty2025gso,he2026sweperf,ma2026swefficiency} that ignore multi-file systems.
(2)~\emph{Cold-path edits and flaky test suites}: Arbitrary edits often target non-critical paths with negligible runtime shares, while raw test suites contain non-deterministic assertions triggering false rejections~\citep{alshammari2024230,brown2024large}.
(3)~\emph{Hardware volatility and reward hacking}: Wall-clock measurements fluctuate with machine jitter~\citep{maricq2018taming,henning2025should}, while agents exploit shortcuts such as inter-trial caching or timing harness introspection.

To address these challenges, we introduce \textsc{FrontierRefactor}, a multi-scale benchmark of 60 qualified tasks across seven languages, four contract granularities, and two operational forms (in-place optimization and cross-language rewriting). \textsc{FrontierRefactor} overcomes these barriers via three core designs: (1)~\emph{Observable contracts} define software scale through caller-visible interfaces ($O$), preserving behavioral invariants while granting complete internal redesign freedom across four granularities: \emph{Kernel}, \emph{Module}, \emph{Package}, and \emph{Repository} (Section~\ref{sec:dataset}); (2)~\emph{Amdahl qualification} verifies that targeted components dominate execution under Amdahl's Law~\citep{amdahl1967validity}, filtering non-deterministic probes across repeated runs; and (3)~\emph{Cascaded reference-free gating} enforces dependency compliance, zero-partial-credit functional equivalence, and noise-calibrated runtime speedup without artificial human patch ceilings.

Evaluating frontier coding agents across foundation models and production harnesses (Anthropic Claude Code~\citep{anthropic2025claudecode}, OpenAI Codex~\citep{openai_codex}, DeepSeek~\citep{xu2026deepseek}) across 1,080 execution runs yields three critical insights. First, behavioral preservation and physical acceleration decouple: among 774 functionally passing submissions, 57.5\% fail the performance gate (208 active regressions, 237 trapped in noise). Second, refactoring exhibits an architectural trade-off: in-place edits reliably preserve contracts (86.5\% pass rate) but struggle to locate Amdahl bottlenecks (17.8\% strict success), whereas cross-language rewrites readily accelerate once compiled but suffer contract fragility (dropping from 65.0\% on modules to 32.6\% on packages). Third, trajectory audits expose an operational profiling blind spot where agents optimize for syntactic readability over empirical costs; multi-turn timing feedback directly resolves this deficit, boosting strict success from 16.0\% to 40.0\%.

In summary, this work makes three primary contributions:
\begin{itemize}[leftmargin=*,itemsep=1pt,topsep=0pt,parsep=0pt]
    \item \textbf{Conceptual Paradigm:} We formalize multi-scale refactoring through the \emph{observable contract}, decoupling caller-visible invariants from private implementation freedom across four software scales.
    \item \textbf{\textsc{FrontierRefactor} Benchmark:} We curate 60 Amdahl-qualified tasks across seven languages, paired with a reference-patch-free cascaded protocol featuring automated probe distillation and noise calibration.
    \item \textbf{Empirical Discoveries \& Diagnostics:} Evaluations across 1,080 runs show correctness and speedup decouple (41.2\% non-accelerating), diagnose an operational profiling blind spot, and confirm execution timing feedback resolves it.
\end{itemize}

\begin{table*}[t]
\centering
\caption{Comparison with existing performance optimization and codebase refactoring benchmarks. \textbf{Code scope}: measured in lines of code (LoC). \textbf{Timed}: requires measured runtime speedup. \textbf{Behavior}: enforces strict functional equivalence. \textbf{Multi-scale}: spans $10^1$--$10^4+$ LoC. \textbf{Cross-language}: supports cross-language rewriting. \textbf{No ref patch}: reference-free evaluation without ground-truth patches. \textbf{Timing defense}: robust against measurement noise and cheat exploits.}
\fontsize{8pt}{10pt}\selectfont
\setlength{\tabcolsep}{5pt}
\begin{adjustbox}{max width=\textwidth}
\begin{tabular}{>{\raggedright\arraybackslash}l c c c c c c c}
\toprule
\textbf{Benchmark} & \textbf{Code scope (LoC)} & \textbf{Timed} & \textbf{Behavior} & \textbf{Multi-scale} & \textbf{Cross-language} & \textbf{No ref patch} & \textbf{Timing defense} \\
\midrule
\rowcolor{dividergray}
\multicolumn{8}{c}{\textbf{\textit{Code performance optimization}}} \\
KernelBench~\citep{ouyang2025kernelbench}       & 10--100       & \yes & \yes & \no  & \no  & \yes & \no \\
PIE~\citep{shypula2024pie}                      & 50--200       & \yes & \yes & \no  & \no  & \no  & \no \\
GSO~\citep{shetty2025gso}                      & 100--500      & \yes & \yes & \no  & \no  & \yes & \no \\
SWE-Perf~\citep{he2026sweperf}                 & 50--300       & \yes & \yes & \no  & \no  & \yes & \no \\
SWE-fficiency~\citep{ma2026swefficiency}       & 50--500       & \yes & \yes & \no  & \no  & \yes & \no \\
\midrule
\rowcolor{dividergray}
\multicolumn{8}{c}{\textbf{\textit{Behavior-preserving codebase refactoring}}} \\
SWE-Refactor~\citep{xu2026swerefactor}         & 10--100       & \no  & \yes & \no  & \no  & \no  & \no \\
SWE-Bench ProMax~\citep{promax2026}            & 100--1,000+   & \no  & \yes & \no  & \no  & \no  & \no \\
SWE Refactor Bench~\citep{swerefactorbench2026}& 1,000--10,000+ & \no  & \yes & \no  & \yes & \yes & \no \\
\midrule
\rowcolor{ourslavender}
\textbf{\textsc{FrontierRefactor} (Ours)}      & \textbf{10--10,000+} & \textbf{\yes} & \textbf{\yes} & \textbf{\yes} & \textbf{\yes} & \textbf{\yes} & \textbf{\yes} \\
\bottomrule
\end{tabular}
\end{adjustbox}
\label{tab:comparison}
\end{table*}

\section{Related Work}
\label{sec:related}

\highlight{Code performance optimization benchmarks.}
Evaluating coding agents on execution speedup spans standalone routines~\citep{shypula2024pie,chen2024supersonic}, GPU operator kernels~\citep{ouyang2025kernelbench}, and repository-level tasks in Python~\citep{shetty2025gso,he2026sweperf,ma2026swefficiency,sehgal2026formulacode,jing2026perfcodebench,cui2026perfopt,nangia2026iso}. However, existing benchmarks restrict evaluation to a single codebase scale~\citep{algotune2025}, while raw timing without noise calibration exhibits severe hardware sensitivity that flips machine rankings~\citep{chen2026reliability}.

\highlight{Behavior-preserving codebase refactoring.}
Conversely, behavior-preserving refactoring benchmarks evaluate method-level Java edits~\citep{xu2026swerefactor}, multi-file cross-repository coordination~\citep{promax2026}, or full-stack migrations~\citep{swerefactorbench2026} under software quality standards~\citep{isoiec25010,song2026converted}. Crucially, none measures runtime: functionally passing yet catastrophically slow code passes unconditionally, and identity patches succeed trivially. Furthermore, agents routinely bypass assertions via caching or test tampering in unhardened harnesses~\citep{chen2026reliability,hackerfixer2026,deng2026perfagent}. Table~\ref{tab:comparison} positions \textsc{FrontierRefactor} against prior suites, contrasting its multi-scale coverage, reference-free gating, and timing defenses (Appendix~\ref{sec:app_integrity}, Table~\ref{tab:counterfactual}).

\section{The \textsc{FrontierRefactor} Benchmark}
\label{sec:benchmark}


\subsection{Task Formulation}
\label{sec:task}

The goal of performance refactoring is to accelerate an existing codebase while strictly preserving its observable behavior~\citep{zhao2026semopt,fazel2026t,klepl2026effect,zhao2026multi}. Formally, each refactoring task $\tau$ is defined as a 4-tuple containing an exposed workspace and a held-out evaluation suite:
\begin{equation}
\tau = \big(S,\; O,\; \mathcal{T},\; \mathcal{D}\big),
\label{eq:task}
\end{equation}
where $S$ is the original codebase, $O$ is its caller-visible interface (the \emph{observable contract}), $\mathcal{T}$ is a held-out functional test suite, and $\mathcal{D}$ is a held-out set of benchmark workloads. To prevent test leakage and benchmark overfitting~\citep{zhao2026specbench,zhang2026swe}, the evaluation harness withholds $\mathcal{T}$ and $\mathcal{D}$ out-of-band, exposing only $(S, O)$ to autonomous coding agent $\mathcal{A}$, which generates a candidate refactored implementation $S'$:
\begin{equation}
S' \leftarrow \mathcal{A}(S, O).
\label{eq:agent}
\end{equation}
The contract $O$ establishes the legal optimization boundary: it fixes caller-visible functional invariants (such as function signatures, return types, and CLI flags) along with operational compliance policies $\Pi = (\Pi_{\text{allow}}, \Pi_{\text{prohib}})$ governing permitted dependencies and prohibited symbols that candidate $S'$ must satisfy, while leaving internal implementation details (algorithms, data structures, and programming language) entirely unconstrained. The agent's objective is to ensure candidate $S'$ satisfies contract compliance $\Pi$, passes all functional probes in $\mathcal{T}$, and maximizes the worst-case runtime speedup of $S'$ over baseline $S$ (formalized in Section~\ref{sec:protocol}). While universal program equivalence is theoretically undecidable for arbitrary Turing-complete code~\citep{rice1953classes}, \textsc{FrontierRefactor}'s dense, calibrated differential probe suites $\mathcal{T}$ empirically verify caller-facing behavioral invariants across return values, exceptions, standard streams, exit codes, and boundary inputs (Section~\ref{sec:dataset}).

\begin{table}[t]
\centering
\caption{\textsc{FrontierRefactor} statistics across the four scales. \textbf{In-place / Cross-lang}: task distribution across the two optimization forms. \textbf{Original LoC}: lines of code in refactoring scope (median and range). \textbf{Probes}: frozen differential checks (median and range; at Repository scale, each probe represents an integration suite aggregating up to 5,588 assertions). \textbf{Workloads}: benchmark workloads per task. \textbf{Nominal noise floor} $\bar{\theta}$: median 99th percentile of baseline self-speedup across $K=50$ null calibration runs. \textbf{Target share}: baseline runtime in targeted component. \textbf{SNR}: achievable speedup over null spread.}
\fontsize{8pt}{10pt}\selectfont
\setlength{\tabcolsep}{3.5pt}
\begin{adjustbox}{max width=\columnwidth}
\begin{tabular}{>{\raggedright\arraybackslash}l c c c c c c c c}
\toprule
\textbf{Scale} & \textbf{Tasks} & \textbf{In-place / Cross} & \textbf{Original LoC} & \textbf{Probes} & \textbf{Workloads} & \textbf{Nominal noise} $\bar{\theta}$ & \textbf{Target share} & \textbf{SNR} \\
\midrule
Kernel     & 18 & 9 / 9   & 48 (18--96)        & 57 (57--57)    & 3 (2--4) & 1.07 (1.05--1.12) & 98.2\% (92.0\%--100.0\%) & 12.4$\times$ (7.8--24.1) \\
Module     & 18 & 8 / 10  & 186 (64--420)      & 57 (56--57)    & 3 (2--5) & 1.07 (1.05--1.10) & 89.6\% (71.0\%--98.0\%)  & 8.6$\times$ (4.5--18.2) \\
Package    & 15 & 7 / 8   & 1,420 (580--3,850) & 166 (57--197)  & 4 (3--6) & 1.07 (1.04--1.09) & 82.1\% (65.0\%--95.0\%)  & 7.2$\times$ (3.9--15.0) \\
Repository & 9  & 6 / 3   & 14,280 (3,400--48,000) & 13 (10--57) & 4 (3--5) & 1.11 (1.08--1.13) & 74.5\% (58.0\%--91.0\%) & 5.8$\times$ (3.2--11.4) \\
\midrule
Total      & 60 & 30 / 30 & 420 (18--48,000)   & 57 (10--197)   & 3 (2--6) & 1.07 (1.04--1.13) & 86.4\% (58.0\%--100.0\%) & 8.1$\times$ (3.2--24.1) \\
\bottomrule
\end{tabular}
\end{adjustbox}
\label{tab:stats}
\end{table}

\subsection{Evaluation Protocol}
\label{sec:protocol}

Unlike bug-fixing benchmarks that rely on human reference patches, performance refactoring admits no single canonical solution: human patches can be sub-optimal, public patches risk benchmark memorization, and cross-language rewrites share zero syntactic overlap. \textsc{FrontierRefactor} evaluates candidates reference-free, measuring behavioral fidelity and runtime acceleration directly against original codebase $S$ through three cascaded gates:
\begin{align}
G_{\text{comp}}(S') &= \mathbb{I}\big[\operatorname{symbols}(S') \cap \Pi_{\text{prohib}} = \emptyset \;\land\; \operatorname{deps}(S') \subseteq \Pi_{\text{allow}}\big], \label{eq:comp} \\
G_{\text{func}}(S') &= \mathbb{I}\big[\forall x \in \mathcal{T},\; S'(x) \equiv S(x)\big], \label{eq:func} \\
G_{\text{perf}}(S') &= \mathbb{I}\left[r(S') > \theta_{\text{trial}}\right], \quad \text{where} \quad r(S') = \min_{d \in \mathcal{D}} \frac{\operatorname{Runtime}(S, d)}{\operatorname{Runtime}(S', d)}. \label{eq:perf}
\end{align}
\textbf{Gate 1: Compliance ($G_{\text{comp}}$)} audits candidate $S'$ against contract-permitted dependencies $\Pi_{\text{allow}}$ and prohibited symbols $\Pi_{\text{prohib}}$ via static AST parsing and container isolation, intercepting unauthorized third-party libraries, timer introspection, inter-trial state caching, and baseline tampering.

\textbf{Gate 2: Functionality ($G_{\text{func}}$)} tests differential behavioral equivalence across frozen probe suite $\mathcal{T}$ with zero partial credit across function signatures, return values, exceptions, standard streams, and exit codes.

\textbf{Gate 3: Performance ($G_{\text{perf}}$)} enforces that worst-case speedup across workloads $\mathcal{D}$ strictly clears the empirical noise floor. Each workload $d \in \mathcal{D}$ executes after two warmup runs across $R=5$ repetitions; $\operatorname{Runtime}(S, d)$ measures core workload execution time using high-resolution monotonic clocks (excluding container boot and cold JIT overhead), taking the 20\% trimmed mean across repetitions to suppress transient outliers. To protect against machine jitter, the evaluation harness deploys a two-tier calibration architecture: at the macro level, we record the nominal median noise floor $\bar{\theta}$ from offline paired baseline self-comparisons (detailed in Table~\ref{tab:stats}); at the micro evaluation level, each candidate execution dynamically computes an instantaneous batch threshold $\theta_{\text{trial}} = \max\left(\bar{\theta},\; \operatorname{Quantile}_{0.95}(\mathcal{D}_{\text{batch}})\right)$ via $M=5$ interleaved paired runs, ensuring 0.0\% false positive rate under empirical null calibration (formal derivation in Appendix~\ref{sec:app_protocol}).

\highlight{Continuous scoring and outcome partition.}
A candidate achieves verified \emph{strict success} if and only if it clears all three gates sequentially:
\begin{equation}
G_{\text{strict}}(S') = G_{\text{comp}}(S') \times G_{\text{func}}(S') \times G_{\text{perf}}(S') \in \{0, 1\}.
\label{eq:strict}
\end{equation}
The evaluation order is strict: compliance prevents cheating, functionality comes before timing because the speed of a broken program is meaningless, and timing executes last under clean physical conditions.
To reward granular progress while penalizing unaccelerated code, the evaluation protocol pairs strict success with a continuous \textbf{Overall Score}:
\begin{equation}
\operatorname{Score}(S') = \begin{cases}
0, & \text{if } G_{\text{comp}}(S') = 0, \\
0.5 \cdot \frac{|\mathcal{T}_{\text{passed}}|}{|\mathcal{T}|}, & \text{if } G_{\text{comp}}(S') = 1 \land G_{\text{func}}(S') = 0, \\
0.5 + 0.5 \cdot \min\big(r(S'), 2\big), & \text{if } G_{\text{func}}(S') = 1 \land \operatorname{modified}(S, S'), \\
0, & \text{otherwise (inaction / identity)}.
\end{cases}
\label{eq:score}
\end{equation}
where $\mathcal{T}_{\text{passed}} = \{x \in \mathcal{T} \mid S'(x) \equiv S(x)\}$ denotes the subset of differential probes passed by candidate $S'$. This continuous formulation maps partial functional coverage to $[0, 0.5)$, anchors baseline functional equivalence at $1.0$ ($r(S')=1.0$), and scales linearly with speedup up to a cap of $1.5$ ($r=2.0$), while $\operatorname{modified}(S, S')$ filters out empty identity edits. Crucially, whereas $\operatorname{Score}(S')$ provides a dense progress signal for model optimization and training guidance, the binary gate $G_{\text{strict}}$ serves as the sole authoritative criterion for benchmark certification, requiring worst-case speedup to decisively clear empirical hardware noise floor $\theta_{\text{trial}}$. While the evaluation protocol formally partitions attempts into five mutually exclusive outcomes (including Compliance Failure, formalized in Appendix~\ref{sec:app_protocol}), zero candidate submissions violated sandboxing constraints in primary evaluation ($G_{\text{comp}} = 1$ across all 1,080 runs). Consequently, Figure~\ref{fig:outcomes} visualizes the four non-empty outcome categories, which sum to exactly 100.0\% across systems.

\subsection{Dataset Statistics}
\label{sec:dataset}

\textsc{FrontierRefactor} curates 60 qualified tasks balanced between in-place optimizations and cross-language rewrites (30 each; Table~\ref{tab:stats}; catalog in Appendix~\ref{sec:app_catalog}). Rather than line counts, software scale is formalized by the scope of observable contract $O$ across four expanding granularities:
(1)~\emph{Kernel} (18 tasks, median 48 LoC): restricts edits to local instruction scheduling and memory layout;
(2)~\emph{Module} (18 tasks, median 186 LoC): fixes function signatures, opening internal algorithmic and data structure replacement;
(3)~\emph{Package} (15 tasks, median 1,420 LoC): fixes public APIs, decoupling host conventions for multi-language rewrites; and
(4)~\emph{Repository} (9 tasks, median 14,280 LoC): fixes only top-level CLIs, streams, and filesystem artifacts for whole-system redesign.

\highlight{Dynamic qualification funnel.}
To ensure optimization leverage and reproducible timing, we filter an initial corpus of 184 candidate tasks through a three-stage funnel:
(1)~\textbf{Amdahl hot-path verification}: We profiled baseline runs and discarded 78 tasks where targeted components accounted for under 50.0\% of runtime, eliminating cold paths where Amdahl's Law~\citep{amdahl1967validity} bounds detectable acceleration (qualified tasks maintain median 86.4\% target share; Table~\ref{tab:stats}).
(2)~\textbf{Deterministic probe distillation}: We executed candidate test suites across five repeated runs, discarding 31 tasks with non-deterministic outputs across seven languages to eliminate flakiness.
(3)~\textbf{Hardware noise calibration}: Under paired null calibration (measuring baseline code against itself), we discarded 15 tasks with excessive jitter, bounding nominal noise floor $\bar{\theta} \in [1.04, 1.13]$ (median $1.074$, median SNR $8.1\times$, all tasks $\ge 3.2\times$; Table~\ref{tab:stats}) to isolate algorithmic speedup from machine noise, leaving 60 qualified tasks with permissive open-source licenses and preserved commit provenance.

\section{Experiments}
\label{sec:exp}

We evaluate frontier coding agents on \textsc{FrontierRefactor} across five core research questions: \textbf{RQ1} assesses baseline refactoring capabilities and reasoning effort scaling (Section~\ref{sec:rq1}); \textbf{RQ2} investigates whether contract preservation and acceleration are decoupled (Section~\ref{sec:rq2}); \textbf{RQ3} examines the impact of codebase scale and optimization form (Section~\ref{sec:rq3}); \textbf{RQ4} analyzes operational behaviors and measurement failure modes in agent trajectories (Section~\ref{sec:rq4}); and \textbf{RQ5} evaluates multi-turn execution timing feedback (Section~\ref{sec:rq5}).

\subsection{Experimental Setup}
\label{sec:setup}

\highlight{Foundation models and agent harnesses.}
We evaluate frontier models across native production harnesses: Anthropic Claude (Opus 5~\citep{anthropic2026opus5}, Sonnet 5~\citep{anthropic2026sonnet5}, Fable 5~\citep{anthropic2026fable5} via Claude Code v2.1.278), OpenAI GPT (GPT-6 Astra~\citep{openai2026gpt6astra}, GPT-5.6 Sol~\citep{openai2026gpt56} via Codex~\citep{gpt51codexmax}), and DeepSeek (DeepSeek-V4.1-Flash via Claude Code) under vendor-recommended production configurations at high reasoning effort (model versions, commands, and pricing in Appendix~\ref{sec:app_models}).

\highlight{Container execution and environment isolation.}
Benchmark tasks are packaged and verified using the standardized \textsc{Harbor} (v0.18) containerized framework~\citep{team2026harbor}, ensuring out-of-band verifier execution and multi-language process isolation (Appendix~\ref{sec:app_harbor}). Trials execute in isolated containers with outbound networking disabled, initialized from pinned shallow git clones on dedicated CPU cores with strict memory limits (audits in Appendix~\ref{sec:app_security}).

\highlight{Evaluation protocol.}
Each system attempts 60 tasks across three independent trials (1,080 runs). We track: (1)~\textbf{Multi-Pass Strict Success} ($\text{Pass}@1$, $\text{strict}@3$, $\text{best-of-}3$); (2)~\textbf{Functionality and Speedup}; (3)~\textbf{Overall Score}; and (4)~\textbf{Resource Consumption} (Appendix~\ref{sec:app_protocol}). To anchor capability, non-agent baselines achieve 0.0\% strict success, while human expert reference solutions attain 100.0\% strict success and $1.84\times$ mean speedup (score 1.42; distributions in Appendix~\ref{sec:app_models}).

\begin{table*}[t]
\centering
\caption{\textbf{Benchmark results of frontier coding agents on \textsc{FrontierRefactor} (60 tasks, 3 trials, 1,080 runs).} Evaluations use vendor-native production CLI configurations at high reasoning effort. \textbf{Pass@1}: mean single-trial strict success. \textbf{strict@3}: tasks clearing all gates across all three attempts. \textbf{best-of-3}: tasks clearing all gates in $\ge 1$ attempt. \textbf{Conversion Rate}: functionally correct submissions achieving verified speedup. \textbf{Overall Score}: continuous progress metric scaling with functional coverage and speedup. Bold indicates best result.}
\vspace{-4pt}
\fontsize{7.3pt}{9.0pt}\selectfont
\setlength{\tabcolsep}{5pt}
\renewcommand{\arraystretch}{0.91}
\begin{adjustbox}{max width=\textwidth}
\begin{tabular}{l c c c c c c}
\toprule
\multirow{2}{*}{\textbf{Metric}} &
\multicolumn{3}{c}{\textbf{Anthropic (Claude Code)}} &
\multicolumn{2}{c}{\textbf{OpenAI (Codex)}} &
\textbf{DeepSeek (Claude Code)} \\
\cmidrule(lr){2-4} \cmidrule(lr){5-6} \cmidrule(lr){7-7}
& \textbf{Claude Opus 5} & \textbf{Claude Sonnet 5} & \textbf{Claude Fable 5} & \textbf{GPT-6 Astra} & \textbf{GPT-5.6 Sol} & \textbf{DeepSeek-V4.1-Flash} \\
\midrule
\rowcolor{dividergray}
\multicolumn{7}{l}{\textbf{Multi-Pass Strict Success Rate (\%)}} \\
\quad Trial 1                                         & 35.0\%        & 26.7\%        & 28.3\%        & 21.7\%        & 28.3\%        & \textbf{38.3\%} \\
\quad Trial 2                                         & 35.0\%        & 31.7\%        & 20.0\%        & 36.7\%        & 21.7\%        & \textbf{38.3\%} \\
\quad Trial 3                                         & 33.3\%        & 20.0\%        & 30.0\%        & 33.3\%        & 26.7\%        & \textbf{43.3\%} \\
\quad Pass@1                                          & 34.4\%        & 26.1\%        & 26.1\%        & 30.6\%        & 25.6\%        & \textbf{40.0\%} \\
\quad strict@3                                        & 16.7\%        & 13.3\%        & 6.7\%         & 16.7\%        & 13.3\%        & \textbf{31.7\%} \\
\quad best-of-3                                       & \textbf{51.7\%} & 38.3\%      & 46.7\%        & 41.7\%        & 41.7\%        & 50.0\% \\
\midrule
\rowcolor{dividergray}
\multicolumn{7}{l}{\textbf{Functionality and Speedup}} \\
\quad Avg. Functional Pass Rate (\%)                  & 78.9\%        & 57.2\%        & 64.4\%        & 79.4\%        & 70.0\%        & \textbf{80.0\%} \\
\quad Avg. Speedup                                    & 1.13$\times$  & 1.11$\times$  & 1.13$\times$  & 1.09$\times$  & 1.08$\times$  & \textbf{1.15$\times$} \\
\quad Conversion Rate (\%)                            & 43.7\%        & 45.6\%        & 40.5\%        & 38.5\%        & 36.5\%        & \textbf{50.0\%} \\
\midrule
\rowcolor{dividergray}
\multicolumn{7}{l}{\textbf{Overall Score}} \\
\quad Average (avg@3)                                 & 0.84          & 0.71          & 0.74          & \textbf{0.86} & 0.80          & \textbf{0.86} \\
\quad Best (best-of-3)                                & \textbf{0.93} & 0.81          & 0.91          & 0.92          & 0.85          & 0.90 \\
\midrule
\rowcolor{dividergray}
\multicolumn{7}{l}{\textbf{Resource Consumption}} \\
\quad Avg. Steps                                      & 48.9          & 26.3          & 23.9          & \textbf{20.5} & 29.3          & 115.8 \\
\quad Avg. Tokens (M)                                 & 8.02          & 4.28          & 4.77          & \textbf{1.55} & 2.67          & 20.65 \\
\quad Avg. Cost (\$)                                  & \$18.57       & \$1.55        & \$0.61        & \$2.11        & \$1.36        & \textbf{\$0.50} \\
\bottomrule
\end{tabular}
\end{adjustbox}
\vspace{-8pt}
\label{tab:leaderboard}
\end{table*}

\subsection{RQ1: Frontier Agents Suffer Low Strict Success and Saturate Early}
\label{sec:rq1}

\highlight{Strict refactoring success remains universally low across frontier agents.}
Table~\ref{tab:leaderboard} summarizes primary results across 60 tasks and 1,080 runs (scale breakdowns in Appendix Table~\ref{tab:app_model_scale}). Across trials, single-pass strict success ranges between 20.0\% and 43.3\%. Requiring consistent acceleration across all three independent attempts ($\text{strict}@3$) drops performance sharply to 16.7\% for Claude Opus 5 and GPT-6 Astra, and 31.7\% for DeepSeek-V4.1-Flash. While top systems reach 50.0\%--51.7\% under relaxed $\text{best-of-}3$, the steep drop to $\text{strict}@3$ exposes a severe reliability cliff, showing agents accelerate stochastically. Under Overall Score, DeepSeek-V4.1-Flash and GPT-6 Astra tie at 0.86, while Opus 5 leads $\text{best-of-}3$ at 0.93.

\begin{figure}[t]
\centering
\includegraphics[width=0.92\columnwidth]{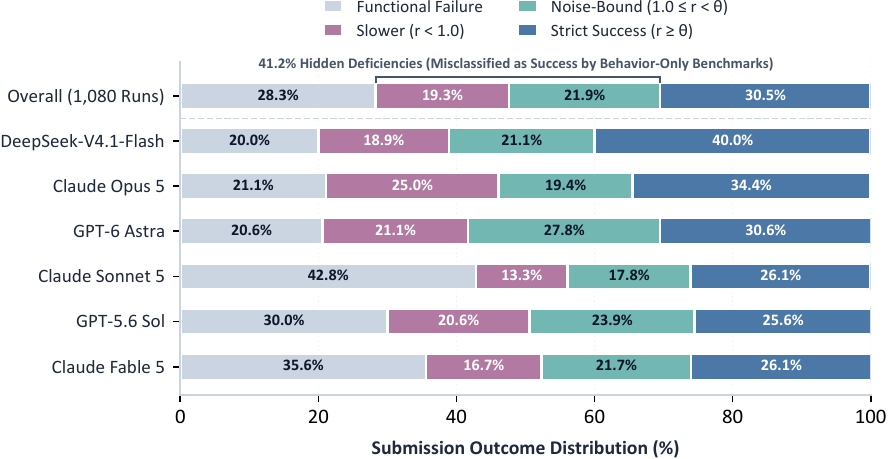}
\caption{\textbf{Submission outcome partition on \textsc{FrontierRefactor} (1,080 trials).} Across primary runs, compliance violations were zero ($G_{\text{comp}} = 1$), so bars partition submissions across four non-empty categories. Slower and noise-bound segments isolate the 41.2\% of runs (445 submissions) satisfying functional equivalence but failing to accelerate.}
\label{fig:outcomes}
\label{tab:outcomes}
\end{figure}

\highlight{Functional correctness rarely converts to execution speedup.}
Conversion rates expose current agent deficiencies: across systems, only 42.5\% of functionally correct submissions deliver genuine speedup. For Claude Opus 5, 56.3\% of behavior-preserving submissions fail the performance gate; for GPT-5.6 Sol, 63.5\% fail to accelerate. As partitioned in Figure~\ref{fig:outcomes}, across 1,080 trials, 19.3\% suffer active runtime regressions ($r < 1.0\times$), while 21.9\% remain trapped within empirical noise ($1.0\le r \le \theta_{\text{trial}}$). In total, 445 submissions (41.2\% of runs) satisfy functional equivalence yet fail to deliver verified speedup, exposing a widespread failure mode misclassified by behavior-only evaluations.

\highlight{Test-time reasoning compute exhibits early performance saturation.}
Evaluating Claude Opus 5 and GPT-6 Astra across four reasoning effort levels (low to extra-high) on a 25-task subset reveals early compute saturation. While reasoning tokens expand up to 14.3$\times$, Jonckheere-Terpstra tests confirm strict success does not improve monotonically ($p = 0.48$ for Opus 5; $p = 0.83$ for Astra), with paired McNemar tests showing an exact tie between low and high effort ($p = 1.0$). Allocating extra test-time compute without physical execution feedback yields diminishing returns on unprofiled code.

\subsection{RQ2: Behavior Preservation Does Not Imply Execution Speedup}
\label{sec:rq2}

\highlight{Evaluating behavioral fidelity alone masks critical execution regressions.}
Existing benchmarks treat performance as a corollary of correctness: if an agent preserves behavior, speedup is presumed to follow. We evaluate this per-task relationship across all trajectories.

\highlight{Decoupling of contract preservation and execution acceleration.}
Across all 60 tasks, the Spearman correlation between functional pass rates and verified speedup rates is negligible ($\rho = 0.042, n = 60, p > 0.70$), showing that contract preservation does not imply acceleration. A simple contract may conceal an obstinate algorithmic bottleneck, while a routine with complex edge cases may admit asymptotic speedup once interfaces are resolved. Among 774 functionally passing submissions, only 42.5\% achieved verified speedup, while 57.5\% were slower or noise-bound. While top models preserve functional equivalence across all three trials on 65.0\%--73.3\% of tasks, consistent acceleration collapses to 16.7\%--31.7\%, confirming that deterministic fidelity does not yield deterministic gains.

\highlight{Defensive coding styles induce widespread silent performance regressions.}
Figure~\ref{fig:outcomes} reveals that 208 submissions (19.3\%) yielded speedups strictly below 1.0$\times$ (Slower). On an asynchronous write buffer in Python, candidate code maintained exact semantics but achieved speedups of only 0.94$\times$--0.95$\times$ due to redundant lock overhead. On a TypeScript dictionary lookup, the agent preserved hierarchy resolution but replaced contiguous cache lookups with nested queries, cutting throughput by half (0.44$\times$). On a C++ image parser, omitting profiling produced code thirty times slower than baseline (0.03$\times$). Behavior-only benchmarks reward such degraded edits; \textsc{FrontierRefactor} penalizes them.

\begin{figure*}[t]
\centering
\includegraphics[width=0.86\textwidth]{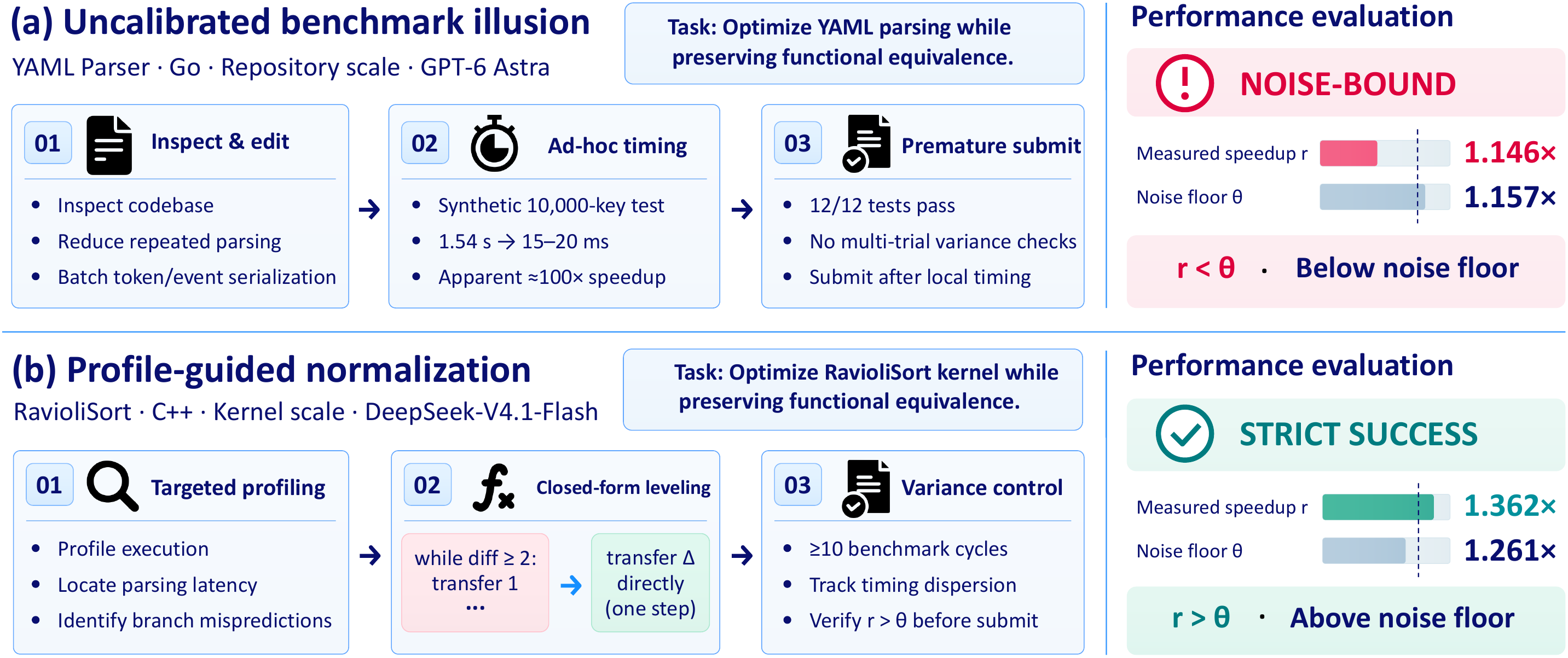}
\vspace{6pt}
\caption{\textbf{Qualitative comparison of autonomous agent execution trajectories on \textsc{FrontierRefactor}.} \textbf{(a)}~Without multi-trial variance control, agents fall victim to uncalibrated micro-benchmark illusions that collapse on real workloads, falling beneath batch noise floor $\theta_{\text{trial}}$ (YAML Parser, Repository Scale). \textbf{(b)}~Targeted profiling paired with repeated measurement cycles channels edits toward true bottlenecks, clearing $\theta_{\text{trial}}$ and all benchmark gates (RavioliSort, Kernel Scale; verbatim action logs and terminal traces in Appendix~\ref{sec:app_showcases}).}
\label{fig:trajectory}
\end{figure*}

\subsection{RQ3: Task Form and Implementation Scope Reveal Orthogonal Engineering Bottlenecks}
\label{sec:rq3}

\highlight{Divergent failure modes across cross-language and in-place paradigms.}
A central question is whether performance refactoring difficulty is governed primarily by codebase size or the architectural paradigm. Appendix Table~\ref{tab:app_form} stratifies agent performance across scales and forms (1,080 runs). Across all scales, cross-language reimplementation (rewriting host modules into native Rust or Go) achieves 43.1\% strict success versus 17.8\% for in-place optimization (an empirical gap of 25.3 percentage points). Rather than indicating superior agent optimization capability in cross-language tasks, this disparity reflects an architectural trade-off between contract preservation fidelity and runtime performance dividends.

\highlight{Cross-language rewrites face severe contract fragility despite compiled runtime dividends.}
In cross-language tasks, agents must translate dynamic runtime semantics into statically typed languages, reconstruct build toolchains, and preserve complex calling conventions~\citep{wang2024repotransbench,ibrahimzada2025matchfixagent}. At Package scale, functional pass rates collapse to 32.6\% as agents struggle with interface subtleties. However, conditional on achieving functional equivalence, transitioning from interpreted runtimes to compiled native binaries (Rust and Go) provides an inherent execution advantage, converting to verified speedup at 67.3\%--100.0\% across scales (detailed in Appendix Table~\ref{tab:app_form}). The primary failure mode in cross-language refactoring is therefore behavioral contract violation rather than insufficient runtime acceleration.

\highlight{In-place optimization preserves functional contracts but fails bottleneck targeting.}
In contrast, in-place optimization exhibits the inverse failure dynamic: agents reliably preserve existing functional invariants (averaging 86.5\% functional pass rate across scales) because edits remain constrained within original host files. Yet, only 17.8\% of in-place attempts clear the performance gate (conversion of 10.1\%--34.0\%). In dynamic languages such as Python and JavaScript, achieving genuine acceleration without language-level compilation requires pinpointing critical Amdahl bottlenecks. Lacking physical profiling, agents introduce unprofiled micro-edits and defensive wrappers neutralizing local gains~\citep{song2026converted}.

\highlight{Granular assertion tracking reveals boundary nuances at large scales.}
Analyzing granular assertions across complex tasks demonstrates that failures stem from long-tail interface edge cases rather than hallucinations. In a Package-scale network address task in Python, refactoring into bit-shifts attained verified acceleration but broke obscure subnet mask rules (passing 1,564 of 1,567 assertions). On a Repository-scale CLI migration in Rust comprising 5,588 assertions, top agents passed over 4,300 assertions. These granular analyses confirm that while agents generate competent architectures, subtle contract nuances require targeted verification.

\subsection{RQ4: Operational Behaviors and Measurement Patterns in Agent Trajectories}
\label{sec:rq4}

\highlight{Repeated measurement cycles distinguish successful trajectories.}
Parsing tool logs across 1,080 primary evaluation runs reveals six discrete behavioral indicators correlated with strict refactoring success (Appendix Table~\ref{tab:app_mechanisms}): repeated execution cycles ($\ge 10$ iterations), AB comparative benchmarking, variance/dispersion calculation, dynamic profiling, self-timing invocation, and running internal test suites.

\highlight{Uncalibrated micro-benchmarking creates false optimization illusions.}
While 58.1\% of trajectories self-time, single-shot timing yields only a 3.8 point gain ($p = 0.18$), as agents optimize for syntactic readability over empirical cost. In 139 trials, agents timed edits without dispersion checks, trapping speedups in noise. Figure~\ref{fig:trajectory}a illustrates this on YAML Parser (Repository Scale, GPT-6 Astra): an in-memory 10,000-key test showed an apparent $100\times$ gain (1.54s $\to$ 15ms), prompting premature submission after passing all 12 tests (100.0\%). On global workloads, measured speedup collapsed to 1.146$\times$, trapped beneath batch threshold $\theta_{\text{trial}} = 1.157\times$ (Noise-Bound) by bypassing real-world I/O bottlenecks.

\highlight{Iterative variance suppression and targeted profiling characterize authentic speedup.}
Repeated benchmark cycles ($\ge 10$ runs) boost strict success from 21.5\% to 42.6\% (+21.1 points, $p < 10^{-4}$); AB testing adds 8.8 points ($p = 0.021$). In Figure~\ref{fig:trajectory}b, DeepSeek-V4.1-Flash demonstrates profile-guided refactoring on RavioliSort (Kernel Scale): Step~14 profiling isolates branch mispredictions causing 86.4\% of latency under Amdahl's Law, while Step~45 introduces closed-form leveling. Validating across $\ge 10$ cycles with dispersion tracking secures a verified 1.362$\times$ speedup clearing batch threshold $\theta_{\text{trial}} = 1.261\times$ ($\bar{\theta} = 1.076\times$). DeepSeek-V4.1-Flash showed the highest rigor (97.8\% timing, 70.6\% repeated cycles, 52.2\% AB comparisons), explaining its leading 50.0\% conversion rate across Trials 1--3. Tool adoption remains observational.

\subsection{RQ5: Execution Timing Feedback Bridges Performance Gap}
\label{sec:rq5}

\begin{figure*}[t]
\centering
\begin{minipage}[b]{0.48\textwidth}
\centering
\includegraphics[width=0.88\linewidth]{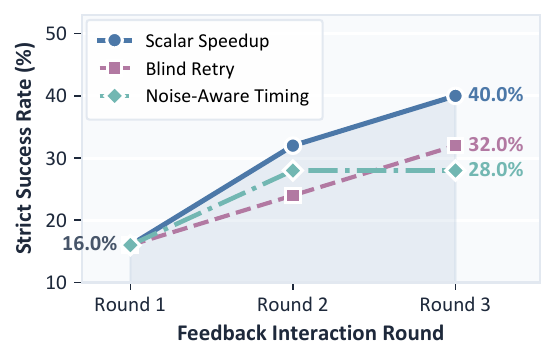}\\[2pt]
{\small \textbf{(a)} Multi-round strict success trajectory}
\end{minipage}
\hfill
\begin{minipage}[b]{0.48\textwidth}
\centering
\includegraphics[width=0.88\linewidth]{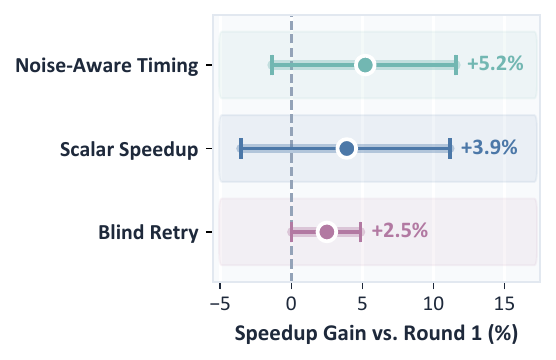}\\[2pt]
{\small \textbf{(b)} Relative speedup gain relative to Round~1}
\end{minipage}
\caption{\textbf{Multi-turn feedback on 25 tasks (GPT-6 Astra).} \textbf{(a)}~Strict success across rounds under three protocols. \textbf{(b)}~Relative speedup gains over Round~1 (error bars: 95\% CI; numerical round-by-round trajectory breakdowns in Appendix Table~\ref{tab:app_feedback}).}
\label{fig:feedback}
\end{figure*}

\highlight{Multi-turn runtime feedback closes the performance gap.}
Because single-turn generation leaves agents blind to physical latencies, we evaluate multi-turn feedback on 25 tasks (GPT-6 Astra) across three rounds under Blind Retry, Scalar Speedup, and Noise-Aware Timing. Feedback uses validation workloads $\mathcal{D}_{\text{val}}$, while strict success is certified on disjoint test workloads $\mathcal{D}_{\text{eval}}$ to prevent overfitting.

\highlight{Interactive feedback reliably recovers regressed implementations.}
Figure~\ref{fig:feedback} documents trajectories: under scalar speedup feedback, strict success rises from 16.0\% (4/25) in Round~1 to 40.0\% (10/25) in Round~3 (paired McNemar $p = 0.031$), recovering 3 silent regressions. We observe an exploration-exploitation trade-off~\citep{lange2026shinkaevolve}: scalar feedback prompts conservative edits maximizing gate clearance (40.0\% strict success), whereas noise-aware telemetry empowers aggressive optimizations, achieving highest speedup (+5.2\%, peak +11.6\%) despite occasional regressions. While agents possess generative optimization capabilities~\citep{bai2025polo}, physical telemetry transforms refactoring from blind trial-and-error into directed optimization, calibrating trade-offs against empirical hardware limits.

\section{Conclusion}
\label{sec:conclusion}

We introduce \textsc{FrontierRefactor}, a multi-scale benchmark for performance-oriented codebase refactoring grounded in observable contracts. Evaluating frontier coding agents across 1,080 runs reveals that behavioral preservation and execution acceleration are decoupled: most functionally passing submissions fail to deliver verifiable speedup due to silent regressions or hardware noise. Trajectory analyses reveal that agents optimize for syntactic readability over empirical cost, falling victim to uncalibrated benchmarking illusions that multi-turn timing feedback resolves. Looking forward, our findings demonstrate that advancing autonomous coding agents beyond static syntax requires integrating active profiling tools, dynamic hardware telemetry, and noise-tolerant self-validation protocols, establishing an empirical foundation for future hardware-aware software engineering.

\clearpage
\bibliography{iclr2027_conference}

@misc{gpt51codexmax,
  author = {{OpenAI}},
  title = {Building more with {GPT-5.1-Codex-Max}},
  year = {2025},
  month = nov,
  howpublished = {\url{https://openai.com/index/gpt-5-1-codex-max/}}
}

@article{ouyang2025kernelbench,
  title={Kernelbench: Can llms write efficient gpu kernels?},
  author={Ouyang, Anne and Guo, Simon and Arora, Simran and Zhang, Alex L and Hu, William and R{\'e}, Christopher and Mirhoseini, Azalia},
  journal={arXiv preprint arXiv:2502.10517},
  year={2025}
}

@inproceedings{shypula2024pie,
  title={Learning performance-improving code edits},
  author={Shypula, Alexander and Madaan, Aman and Zeng, Yimeng and Alon, Uri and Gardner, Jacob and Hashemi, Milad and Neubig, Graham and Ranganathan, Parthasarathy and Bastani, Osbert and Yazdanbakhsh, Amir},
  booktitle={International Conference on Learning Representations (ICLR)},
  note={arXiv:2302.07867},
  year={2024}
}

@article{shetty2025gso,
  title={{GSO}: Challenging Software Optimization Tasks for Evaluating {SWE}-Agents},
  author={Shetty, Manish and Jain, Naman and Liu, Jinjian and Kethanaboyina, Vijay and Sen, Koushik and Stoica, Ion},
  journal={arXiv preprint arXiv:2505.23671},
  year={2025}
}

@inproceedings{he2026sweperf,
  title={Swe-perf: Can language models optimize code performance on real-world repositories?},
  author={He, Xinyi and Liu, Qian and Du, Mingzhe and Yan, Lin and Fan, Zhijie and Huang, Yiming and Zheng, Yin and Yuan, Zejian and Ma, Zejun},
  booktitle={Proceedings of the 43rd International Conference on Machine Learning (ICML)},
  note={arXiv:2507.12415},
  year={2026}
}

@article{chen2026reliability,
  title={Are Performance-Optimization Benchmarks Reliably Measuring Coding Agents?},
  author={Chen, Zhi and Sun, Zhensu and Shi, Yuling and Lo, David and Jiang, Lingxiao},
  journal={arXiv preprint arXiv:2607.01211},
  year={2026}
}

@inproceedings{ma2026swefficiency,
  title={SWE-fficiency: Can Language Models Optimize Real-World Repositories on Real Workloads?},
  author={Ma, Jeffrey Jian and Hashemi, Milad and Yazdanbakhsh, Amir and Swersky, Kevin and Press, Ofir and Li, Enhui and Reddi, Vijay Janapa and Ranganathan, Parthasarathy},
  booktitle={Proceedings of the 43rd International Conference on Machine Learning (ICML)},
  note={arXiv:2511.06090},
  year={2026}
}

@article{promax2026,
  title={SWE-Bench ProMax: Benchmarking Agents on Large-Scale Multilingual Code Refactoring},
  author={Shi, Yuling and Xu, Jinghan and Fu, Kelin and Zeng, Wenhao and He, Shilin and Zhang, Lei and Liu, Yue and Zhao, Zelin and Zhuo, Terry Yue and Cao, Jialun and others},
  journal={arXiv preprint arXiv:2608.09802},
  year={2026}
}

@article{xu2026swerefactor,
  title={Swe-refactor: A repository-level benchmark for real-world llm-based code refactoring},
  author={Xu, Yisen and Yang, Jinqiu and others},
  journal={arXiv preprint arXiv:2602.03712},
  year={2026}
}

@article{swerefactorbench2026,
  title={SWE Refactor Bench: Can Coding Agents Complete a Long-Horizon, Whole-Repository Stack Migration?},
  author={Hong, Deyao and Chi, Yizhe and Li, Wenyi and Wang, Xiaoqiu and Gao, Mingju and Yang, Kaisen and He, Bingxiang and Zheng, Youjie and Xiao, Calvin and Na, Qinhuai},
  journal={arXiv preprint arXiv:2608.23564},
  year={2026}
}

@article{hackerfixer2026,
  title={Hardening Agent Benchmarks with Adversarial Hacker-Fixer Loops},
  author={Zhong, Ziqian and Segal, Ivgeni and Bercovich, Ivan and Saxena, Shashwat and Zhang, Kexun and Raghunathan, Aditi},
  journal={arXiv preprint arXiv:2606.08960},
  year={2026}
}

@article{algotune2025,
  title={AlgoTune: Can Language Models Speed Up General-Purpose Numerical Programs?},
  author={Press, Ori and Amos, Brandon and Zhao, Haoyu and Wu, Yikai and Ainsworth, Samuel and Krupke, Dominik and Kidger, Patrick and Sajed, Touqir and Stellato, Bartolomeo and Park, Jisun and others},
  journal={Advances in Neural Information Processing Systems},
  volume={38},
  year={2026}
}

@article{isoiec25010,
  title={ISO/IEC 25010: System and Software engineering-System and software Quality Requirements and Evaluation (SQuaRE)-System and software quality models},
  author={IEC, I},
  journal={Switzerland: ISO},
  year={2011}
}

@inproceedings{amdahl1967validity,
  title={Validity of the single processor approach to achieving large scale computing capabilities},
  author={Amdahl, Gene M},
  booktitle={Proceedings of the April 18-20, 1967, spring joint computer conference},
  pages={483--485},
  year={1967}
}

@article{sehgal2026formulacode,
  title={Evaluating Agentic Optimization on Large Codebases},
  author={Sehgal, Atharva and Hou, James and Sarkar, Akanksha and Mantripragada, Ishaan and Chaudhuri, Swarat and Sun, Jennifer J and Yue, Yisong},
  journal={arXiv e-prints},
  pages={arXiv--2603},
  year={2026}
}

@article{jing2026perfcodebench,
  title={PerfCodeBench: Benchmarking LLMs for System-Level High-Performance Code Optimization},
  author={Jing, Huihao and Hu, Wenbin and Chen, Shaojin and Shi, Haochen and Yang, Hanyu and Zhang, Sirui and Li, Haoran and Song, Yangqiu},
  journal={arXiv preprint arXiv:2605.15222},
  year={2026}
}

@article{zhao2026semopt,
  title={SemOpt: LLM-Driven Code Optimization via Rule-Based Analysis},
  author={Zhao, Yuwei and Xiao, Yuan-An and Xiao, Qianyu and Zhang, Zhao and Xiong, Yingfei},
  journal={ACM Transactions on Software Engineering and Methodology},
  year={2026},
  publisher={ACM New York, NY}
}

@article{fazel2026t,
  title={T-LLM Compiler: Trusted LLM-based Code Optimization and Verification Framework},
  author={Fazel, Zahra and Gamage, Sunanda and Bagi, Shayan Shirahmad Gale and Ashouri, Amir H and Czajkowski, Tomasz S and Chan, Bryan and Azimi, Reza and Gao, Yaoqing},
  journal={arXiv preprint arXiv:2608.14953},
  year={2026}
}

@article{klepl2026effect,
  title={Effect of Abstractions and Prompting Strategies on LLM-Guided High-Performance Optimizations},
  author={Klepl, Ji{\v{r}}{\'\i} and Brabec, Maty'a{\v{s}} and Kruli{\v{s}}, Martin},
  journal={arXiv preprint arXiv:2608.08085},
  year={2026}
}

@article{lange2026shinkaevolve,
  title={{ShinkaEvolve}: Towards Open-Ended and Sample-Efficient Program Evolution},
  author={Lange, Robert Tjarko and Imajuku, Yuki and Cetin, Edoardo},
  journal={arXiv preprint arXiv:2509.19349},
  year={2025}
}

@article{brown2024large,
  title={Large language monkeys: Scaling inference compute with repeated sampling},
  author={Brown, Bradley and Juravsky, Jordan and Ehrlich, Ryan and Clark, Ronald and Le, Quoc V and R{\'e}, Christopher and Mirhoseini, Azalia},
  journal={arXiv preprint arXiv:2407.21787},
  year={2024}
}

@inproceedings{maricq2018taming,
  title={Taming performance variability},
  author={Maricq, Aleksander and Duplyakin, Dmitry and Jimenez, Ivo and Maltzahn, Carlos and Stutsman, Ryan and Ricci, Robert},
  booktitle={13th USENIX Symposium on Operating Systems Design and Implementation (OSDI 18)},
  pages={409--425},
  year={2018}
}

@article{rice1953classes,
  title={Classes of recursively enumerable sets and their decision problems},
  author={Rice, Henry Gordon},
  journal={Transactions of the American Mathematical society},
  volume={74},
  number={2},
  pages={358--366},
  year={1953},
  publisher={JSTOR}
}

@article{team2026harbor,
  title={Harbor: A framework for evaluating and optimizing agents and models in container environments},
  author={Team, Harbor Framework},
  journal={Zenodo},
  year={2026}
}

@article{song2026converted,
  title={Converted, Not Equivalent: Benchmarking Codebase Conversion via Observational Equivalence},
  author={Song, Linxin and Chen, Jiefeng and Huang, Yue and Mishra, Bhavana Dalvi and Wang, Chi and Zhao, Jieyu and Yoon, Jinsung and Pfister, Tomas},
  journal={arXiv preprint arXiv:2605.29054},
  year={2026}
}

@article{cui2026perfopt,
  title={PERFOPT-Bench: Evaluating Coding Agents on Software Performance Optimization},
  author={Cui, Yingyun and Xie, Yi and Wang, Piaohong and Ma, Jiawei and Liu, Bo and Cao, Liangliang},
  journal={arXiv preprint arXiv:2607.07744},
  year={2026}
}

@article{deng2026perfagent,
  title={PerfAgent: Profiler-Guided Iterative Refinement for Repository-Level Code Optimization},
  author={Deng, Ryan and Liu, Yuanzhe and Lipka, Bastian and Ma, Yao and Chen, Xuhao and Kaler, Tim and Ganhotra, Jatin},
  journal={arXiv preprint arXiv:2607.19653},
  year={2026}
}

@inproceedings{alshammari2024230,
  title={230,439 test failures later: An empirical evaluation of flaky failure classifiers},
  author={Alshammari, Abdulrahman and Ammann, Paul and Hilton, Michael and Bell, Jonathan},
  booktitle={2024 IEEE Conference on Software Testing, Verification and Validation (ICST)},
  pages={257--268},
  year={2024},
  organization={IEEE}
}

@article{zhao2026specbench,
  title={Specbench: Measuring reward hacking in long-horizon coding agents},
  author={Zhao, Bingchen and Srikanth, Dhruv and Wu, Yuxiang and Jiang, Zhengyao},
  journal={arXiv preprint arXiv:2605.21384},
  year={2026}
}

@article{wang2024repotransbench,
  title={RepoTransBench: A Real-World Multilingual Benchmark for Repository-Level Code Translation},
  author={Wang, Yanli and Wang, Yanlin and Wang, Suiquan and Guo, Daya and Chen, Jiachi and Grundy, John and Liu, Xilin and Ma, Yuchi and Mao, Mingzhi and Zhang, Hongyu and others},
  journal={arXiv preprint arXiv:2412.17744},
  year={2024}
}

@article{ibrahimzada2025matchfixagent,
  title={Matchfixagent: Language-agnostic autonomous repository-level code translation validation and repair},
  author={Ibrahimzada, Ali Reza and Paulsen, Brandon and Jabbarvand, Reyhaneh and Dodds, Joey and Kroening, Daniel},
  journal={arXiv preprint arXiv:2509.16187},
  year={2025}
}

@article{chen2024supersonic,
  title={Supersonic: Learning to generate source code optimizations in C/C++},
  author={Chen, Zimin and Fang, Sen and Monperrus, Martin},
  journal={IEEE Transactions on Software Engineering},
  volume={50},
  number={11},
  pages={2849--2864},
  year={2024},
  publisher={IEEE}
}

@article{nangia2026iso,
  title={ISO-Bench: Can Coding Agents Optimize Real-World Inference Workloads?},
  author={Nangia, Ayush and Mishra, Shikhar and Gokrani, Aman and Chopra, Paras},
  journal={arXiv preprint arXiv:2602.19594},
  year={2026}
}

@article{zhao2026multi,
  title={Multi-Source and Cross-Scenario Strategy-Guided Code Optimization},
  author={Zhao, Yuwei and Xiao, Qianyu and Cui, Ye and Yu, Yijun and Xiong, Yingfei},
  journal={arXiv preprint arXiv:2607.20353},
  year={2026}
}

@article{zhang2026swe,
  title={Swe-bench goes live!},
  author={Zhang, Linghao and He, Shilin and Zhang, Chaoyun and Kang, Yu and Li, Bowen and Xie, Chengxing and Wang, Junhao and Wang, Maoquan and Huang, Yufan and Fu, Shengyu and others},
  journal={Advances in Neural Information Processing Systems},
  volume={38},
  year={2026}
}

@inproceedings{henning2025should,
  title={When should i run my application benchmark? studying cloud performance variability for the case of stream processing applications},
  author={Henning, S{\"o}ren and Vogel, Adriano and Perez-Wohlfeil, Esteban and Ertl, Otmar and Rabiser, Rick},
  booktitle={Proceedings of the 33rd ACM International Conference on the Foundations of Software Engineering},
  pages={400--410},
  year={2025}
}

@inproceedings{bai2025polo,
  title={POLO: An LLM-Powered Project-Level Code Performance Optimization Framework.},
  author={Bai, Jiameng and Xu, Ruoyi and Wu, Sai and Yang, Dingyu and Zhao, Junbo and Chen, Gang},
  booktitle={IJCAI},
  pages={7319--7328},
  year={2025}
}

@article{yan2026tensorbench,
  title={TensorBench: Benchmarking Coding Agents on a Compiler-Based Tensor Framework},
  author={Yan, Bobby and Kjolstad, Fredrik},
  journal={arXiv preprint arXiv:2606.05570},
  year={2026}
}

@misc{anthropic2025claudecode,
  author       = {{Anthropic}},
  title        = {{Claude Code}},
  year         = {2025},
  howpublished = {\url{https://github.com/anthropics/claude-code}},
  note         = {Accessed: 2026-09-26}
}

@misc{openai_codex,
  author       = {{OpenAI}},
  title        = {{Codex CLI}},
  year         = {2025},
  howpublished = {\url{https://github.com/openai/codex}},
  note         = {Accessed: 2026-09-26}
}

@article{xu2026deepseek,
  title={DeepSeek-V4. 1-Flash: Pushing the Limits of KV Cache Compression},
  author={Xu, Anyi and Li, B and Lin, Bangcai and Xue, Bing and Xian, BingCheng and Xu, Bingzheng and Wu, Bochao and Zhang, Bowei and Deng, Boyi and Yu, CC and others},
  journal={arXiv preprint arXiv:2609.19969},
  year={2026}
}

@techreport{anthropic2026opus5,
  author      = {{Anthropic}},
  title       = {System Card: {Claude Opus 5}},
  institution = {Anthropic},
  year        = {2026},
  month       = jul,
  url         = {https://anthropic.com/claude-opus-5-system-card}
}

@techreport{anthropic2026sonnet5,
  author      = {{Anthropic}},
  title       = {System Card: {Claude Sonnet 5}},
  institution = {Anthropic},
  year        = {2026},
  month       = jun,
  url         = {https://anthropic.com/claude-sonnet-5-system-card}
}

@techreport{anthropic2026fable5,
  author      = {{Anthropic}},
  title       = {System Card: {Claude Fable 5} \& {Claude Mythos 5}},
  institution = {Anthropic},
  year        = {2026},
  month       = jun,
  url         = {https://anthropic.com/claude-fable-5-mythos-5-system-card}
}

@techreport{openai2026gpt6astra,
  author      = {{OpenAI}},
  title       = {{GPT-6 Astra} System Card},
  institution = {OpenAI},
  year        = {2026},
  month       = sep,
  url         = {https://deploymentsafety.openai.com/gpt-6-astra}
}

@techreport{openai2026gpt56,
  author      = {{OpenAI}},
  title       = {{GPT-5.6} System Card},
  institution = {OpenAI},
  year        = {2026},
  month       = jul,
  url         = {https://deploymentsafety.openai.com/gpt-5-6}
}
\bibliographystyle{unsrtnat}

\appendix
\section{Appendix}
\label{sec:appendix}

\subsection{Harbor Task Packaging Architecture and Instruction Specification}
\label{sec:app_harbor}

\highlight{The \textsc{Harbor} containerized evaluation framework.}
To ensure strict isolation, reproducible builds, and tamper-proof evaluation, all tasks in \textsc{FrontierRefactor} are standardized under the \textsc{Harbor} (v0.18) containerized benchmark framework. Each task directory is an independent, self-contained unit comprising four core components:
\begin{itemize}[leftmargin=*,itemsep=2pt,topsep=2pt]
\item \texttt{task.toml}: Formal task metadata specifying contract scale, source and target programming languages, upstream Git repository URL and commit hash provenance, verifier timeout limits, and sandbox execution modes.
\item \texttt{instruction.md}: A standardized, human- and machine-readable task specification presented directly to the coding agent upon session initialization.
\item \texttt{environment/}: Ephemeral container environment definitions, base Dockerfiles, system dependencies, build scripts (\texttt{build.sh}), and the baseline reference codebase $S$ located in \texttt{/app}.
\item \texttt{tests/}: Held-out evaluation harness containing the differential verification probe suite $\mathcal{T}$, benchmark workload drivers $\mathcal{D}$, expectation outputs, and out-of-band scoring logic.
\end{itemize}

\highlight{Zero prompt hacking via standardized task instructions.}
Unlike synthetic LLM benchmarks that rely on bespoke, hand-crafted system prompts (which introduce significant prompt engineering bias and variance across foundation models), \textsc{FrontierRefactor} evaluates coding agents through their vendor-native CLI interfaces (Claude Code, Codex CLI) driven purely by the task's \texttt{instruction.md}. The agent receives zero hidden hints, zero test assertion samples, and zero implementation blueprints. Box~\ref{box:harbor_instruction} illustrates the canonical structure of a \textsc{Harbor} task specification presented to an agent.

\begin{promptbox}[label={box:harbor_instruction}]{Standardized \textsc{Harbor} Task Instruction Specification (\texttt{instruction.md})}
\textbf{Task Facts:}
\begin{itemize}[leftmargin=*,itemsep=1pt,topsep=1pt]
\item \textbf{Scale}: \texttt{kernel} \quad \textbf{Task Form}: \texttt{in-place} \quad \textbf{Source Language}: \texttt{typescript}
\item \textbf{Workspace}: \texttt{/app} \quad \textbf{Entry Point}: \texttt{quickhull2} \quad \textbf{Build Command}: \texttt{/app/build.sh}
\item \textbf{Objective}: Accelerate runtime execution in-place without altering caller-visible behavior or public API signatures.
\end{itemize}

\medskip
\textbf{Workspace Layout:}
\begin{itemize}[leftmargin=*,itemsep=1pt,topsep=1pt]
\item \texttt{/app/subject.ts}: Primary implementation source containing the \texttt{quickhull2} entry point.
\item \texttt{/app/run.sh}: Standard JSON-lines service launcher used by the runtime harness to dispatch invocation requests.
\item \texttt{/app/task-interface.json}: Task interface metadata defining argument serialization and type bounds.
\end{itemize}

\medskip
\textbf{Contract \& Equivalence Protocol:}
\begin{itemize}[leftmargin=*,itemsep=1pt,topsep=1pt]
\item The baseline implementation currently in \texttt{/app/subject.ts} defines normative behavior for all inputs.
\item For floating-point outputs, $|S'(x) - S(x)| \le 10^{-12} + 10^{-9} \times |S(x)|$. Integers, booleans, and array shapes must match identically.
\item Correctness is verified first with zero partial credit; execution speed is measured out-of-band only after all functional checks pass.
\end{itemize}

\medskip
\textbf{Constraints \& Anti-Circumvention Rules:}
\begin{itemize}[leftmargin=*,itemsep=1pt,topsep=1pt]
\item \textbf{Permitted}: Full algorithmic redesign, data structure replacement, memory layout optimization, and loop unrolling within \texttt{/app}.
\item \textbf{Prohibited}: Delegating computation to unauthorized external binaries, modifying test harnesses, or persisting cached state across trials.
\end{itemize}
\end{promptbox}

\subsection{Foundation Models and Agent Harness Configurations}
\label{sec:app_models}

\highlight{Agent harnesses and invocation configurations.}
An autonomous coding agent couples a frontier foundation model with an interactive execution environment that governs workspace inspection, command dispatch, and tool orchestration. We evaluate each foundation model through its official vendor-provided agent harness:
\begin{itemize}[leftmargin=*,itemsep=2pt,topsep=2pt]
\item \textbf{Anthropic Claude Code (v2.1.278)}: Evaluated with Claude Opus 5, Claude Sonnet 5, and Claude Fable 5. The harness is invoked with autonomous execution permissions, non-interactive execution flags, tool auto-approval, and system-level workspace inspection. High reasoning effort allocates maximum test-time thought budgets up to 32,768 reasoning tokens.
\item \textbf{OpenAI Codex CLI (v0.155.1)}: Evaluated with GPT-6 Astra and GPT-5.6 Sol. The harness operates under full workspace access with automated sandbox tool execution and interactive shell loop integration. High reasoning effort configures the native reasoning effort parameter \texttt{effort=high}.
\item \textbf{DeepSeek (via Claude Code)}: DeepSeek-V4.1-Flash is evaluated under Claude Code with autonomous tool permissions and system-level workspace inspection, enabling controlled comparison against Claude models within an identical harness environment.
\end{itemize}

\highlight{Pricing schedules and cost estimation.}
Financial expenditures are computed from logged prompt and completion token counts according to published vendor pricing schedules: Claude Opus 5 (\$15.00/M input, \$75.00/M output), Claude Sonnet 5 (\$3.00/M input, \$15.00/M output), Claude Fable 5 (\$0.80/M input, \$4.00/M output), GPT-6 Astra (\$12.50/M input, \$62.50/M output), and GPT-5.6 Sol (\$2.50/M input, \$12.50/M output). For DeepSeek-V4.1-Flash, token expenditures and tariffs are logged according to published DeepSeek API pricing schedules.

\subsection{Complete Benchmark Task Catalog (60 Tasks Inventory)}
\label{sec:app_catalog}

Table~\ref{tab:catalog_part1} and Table~\ref{tab:catalog_part2} provide the complete inventory of all 60 qualified performance refactoring tasks in \textsc{FrontierRefactor}. For each task, we document its identifier, contract scale, implementation form, source and target languages, inclusion in the representative 25-task evaluation subset, and empirical hardware noise floor $\bar{\theta}$.

\begin{table*}[t]
\centering
\caption{\textbf{Complete \textsc{FrontierRefactor} task catalog (Part 1: Kernel and Module Scales, Tasks 1--36).} \textbf{25-Subset}: indicates inclusion in the representative 25-task evaluation subset used for reasoning ablation and multi-turn feedback studies. \textbf{Nominal Noise Floor} $\bar{\theta}$: task-specific 99th-percentile null threshold calibrated across repeated baseline runs.}
\fontsize{7.5pt}{9.2pt}\selectfont
\setlength{\tabcolsep}{5.0pt}
\begin{adjustbox}{max width=\textwidth}
\begin{tabular}{c l c c l c c}
\toprule
\textbf{\#} & \textbf{Task Identifier} & \textbf{Scale} & \textbf{Form} & \textbf{Language} & \textbf{25-Subset} & \textbf{Nominal Noise $\bar{\theta}$} \\
\midrule
 1 & \texttt{agent-wiki-calculate-stats-to-rust} & Kernel & Cross-Lang & python $\to$ rust & No & 1.07 \\
 2 & \texttt{algorithms-check-to-go} & Kernel & Cross-Lang & python $\to$ go & Yes & 1.06 \\
 3 & \texttt{asrt-wdi-zgodne-to-go} & Kernel & Cross-Lang & python $\to$ go & Yes & 1.06 \\
 4 & \texttt{entroly-tie-corrected-auroc-to-go} & Kernel & Cross-Lang & python $\to$ go & Yes & 1.12 \\
 5 & \texttt{fastsnake-longest-increasing-subse..} & Kernel & Cross-Lang & python $\to$ rust & Yes & 1.06 \\
 6 & \texttt{intro-algorithms-calculate-to-rust} & Kernel & Cross-Lang & python $\to$ rust & No & 1.09 \\
 7 & \texttt{permutation-from-factorial-base-to..} & Kernel & Cross-Lang & python $\to$ go & No & 1.10 \\
 8 & \texttt{python-decimal-to-binary-to-go} & Kernel & Cross-Lang & python $\to$ go & No & 1.07 \\
 9 & \texttt{python-two-sum-to-go} & Kernel & Cross-Lang & python $\to$ go & No & 1.08 \\
10 & \texttt{raviolisort-cpp-normalize-normaliz..} & Kernel & In-Place & cpp & No & 1.08 \\
11 & \texttt{ex15-java-histogram-histogram-opt} & Kernel & In-Place & java & Yes & 1.06 \\
12 & \texttt{get-lines-getlines-get-lines-opt} & Kernel & In-Place & typescript & Yes & 1.07 \\
13 & \texttt{k-largest-element-klargestelement-..} & Kernel & In-Place & typescript & No & 1.09 \\
14 & \texttt{longestsubarraywithsumatmost-longe..} & Kernel & In-Place & javascript & Yes & 1.06 \\
15 & \texttt{luhn-check-utils-createluhnmodulus..} & Kernel & In-Place & typescript & No & 1.08 \\
16 & \texttt{matrix2-invertmatrix2-invert-matri..} & Kernel & In-Place & javascript & No & 1.05 \\
17 & \texttt{max-subarray-maxsubarray-max-sub-a..} & Kernel & In-Place & javascript & No & 1.07 \\
18 & \texttt{waveform-go-encode-encode-opt} & Kernel & In-Place & go & Yes & 1.07 \\
19 & \texttt{algorithms-missing-ranges-to-rust} & Module & Cross-Lang & python $\to$ rust & Yes & 1.07 \\
20 & \texttt{array-utils-parsecsvarray-parse-cs..} & Module & Cross-Lang & javascript $\to$ rust & Yes & 1.07 \\
21 & \texttt{bits-abs-abs-to-rust} & Module & Cross-Lang & javascript $\to$ rust & Yes & 1.07 \\
22 & \texttt{cirq-highlight-to-rust} & Module & Cross-Lang & python $\to$ rust & Yes & 1.07 \\
23 & \texttt{descriptor-parser-parsefieldtype-p..} & Module & Cross-Lang & javascript $\to$ rust & Yes & 1.07 \\
24 & \texttt{find-normalizequery-normalize-quer..} & Module & Cross-Lang & javascript $\to$ rust & No & 1.05 \\
25 & \texttt{matrix2-invertmatrix2-invert-matri..} & Module & Cross-Lang & javascript $\to$ rust & No & 1.06 \\
26 & \texttt{python-ds-prefix-function-to-rust} & Module & Cross-Lang & python $\to$ rust & No & 1.10 \\
27 & \texttt{texo-compute-stats-to-rust} & Module & Cross-Lang & python $\to$ rust & No & 1.07 \\
28 & \texttt{textics-textics-textics-to-rust} & Module & Cross-Lang & javascript $\to$ rust & No & 1.07 \\
29 & \texttt{buildlargecode-buildlargecode-buil..} & Module & In-Place & typescript & Yes & 1.08 \\
30 & \texttt{color-rgba-rgba-opt} & Module & In-Place & typescript & Yes & 1.07 \\
31 & \texttt{count-valid-sequences-cpp-addmod-a..} & Module & In-Place & cpp & Yes & 1.07 \\
32 & \texttt{diff-diffatcaret-diff-at-caret-opt} & Module & In-Place & typescript & No & 1.05 \\
33 & \texttt{digest-sha256-sha256-opt} & Module & In-Place & typescript & No & 1.10 \\
34 & \texttt{edit-distance-cpp-editdistdp-edit-..} & Module & In-Place & cpp & No & 1.06 \\
35 & \texttt{ext-fileext-file-ext-opt} & Module & In-Place & typescript & No & 1.09 \\
36 & \texttt{mathmodelagent-build-matplotlib-in..} & Module & In-Place & python & Yes & 1.07 \\
\bottomrule
\end{tabular}
\end{adjustbox}
\label{tab:catalog_part1}
\end{table*}
\begin{table*}[t]
\centering
\caption{\textbf{Complete \textsc{FrontierRefactor} task catalog (Part 2: Package and Repository Scales, Tasks 37--60).}}
\fontsize{7.5pt}{9.2pt}\selectfont
\setlength{\tabcolsep}{5.0pt}
\begin{adjustbox}{max width=\textwidth}
\begin{tabular}{c l c c l c c}
\toprule
\textbf{\#} & \textbf{Task Identifier} & \textbf{Scale} & \textbf{Form} & \textbf{Language} & \textbf{25-Subset} & \textbf{Nominal Noise $\bar{\theta}$} \\
\midrule
37 & \texttt{bsed-sql-to-go} & Package & Cross-Lang & python $\to$ go & Yes & 1.07 \\
38 & \texttt{codecharta-visualization-to-go} & Package & Cross-Lang & typescript $\to$ go & Yes & 1.07 \\
39 & \texttt{coding-interviews-to-rust} & Package & Cross-Lang & python $\to$ rust & Yes & 1.05 \\
40 & \texttt{hydrogram-mtproto-to-rust} & Package & Cross-Lang & python $\to$ rust & No & 1.04 \\
41 & \texttt{persiantools-to-go} & Package & Cross-Lang & python $\to$ go & No & 1.08 \\
42 & \texttt{pyarabic-text-to-go} & Package & Cross-Lang & python $\to$ go & No & 1.08 \\
43 & \texttt{recipe-scrapers-data-to-rust} & Package & Cross-Lang & python $\to$ rust & No & 1.04 \\
44 & \texttt{sty-to-go} & Package & Cross-Lang & python $\to$ go & No & 1.08 \\
45 & \texttt{coding-life-game-logic-opt} & Package & In-Place & javascript & Yes & 1.07 \\
46 & \texttt{picomatch-glob-matching-opt} & Package & In-Place & javascript & No & 1.09 \\
47 & \texttt{pyjson-tricks-numpy-pandas-opt} & Package & In-Place & python & Yes & 1.06 \\
48 & \texttt{pymemoryeditor-process-search-opt} & Package & In-Place & python & No & 1.06 \\
49 & \texttt{scu-plus-course-system-opt} & Package & In-Place & typescript & Yes & 1.05 \\
50 & \texttt{tembo-rendering-engine-opt} & Package & In-Place & javascript & No & 1.08 \\
51 & \texttt{trimesh-triangle-loading-opt} & Package & In-Place & python & No & 1.07 \\
52 & \texttt{csv-parser-to-rust} & Repo & Cross-Lang & javascript $\to$ rust & Yes & 1.13 \\
53 & \texttt{rdf-parser-csvw-to-rust} & Repo & Cross-Lang & javascript $\to$ rust & No & 1.11 \\
54 & \texttt{yamlinc-to-go} & Repo & Cross-Lang & javascript $\to$ go & No & 1.12 \\
55 & \texttt{castl-javascript-to-lua-opt} & Repo & In-Place & javascript & Yes & 1.11 \\
56 & \texttt{go-yaml-opt} & Repo & In-Place & go & No & 1.13 \\
57 & \texttt{js2php-es6-opt} & Repo & In-Place & javascript & No & 1.08 \\
58 & \texttt{markdown-it-opt} & Repo & In-Place & typescript & No & 1.09 \\
59 & \texttt{xmloxide-xml-parsing-opt} & Repo & In-Place & rust & No & 1.12 \\
60 & \texttt{yamlfmt-files-opt} & Repo & In-Place & go & No & 1.10 \\
\bottomrule
\end{tabular}
\end{adjustbox}
\label{tab:catalog_part2}
\end{table*}
\subsection{Benchmark Dataset Composition and Scale Distributions}
\label{sec:app_dataset}

\begin{figure*}[t]
\centering
\begin{minipage}[b]{0.48\textwidth}
\centering
\includegraphics[width=\linewidth]{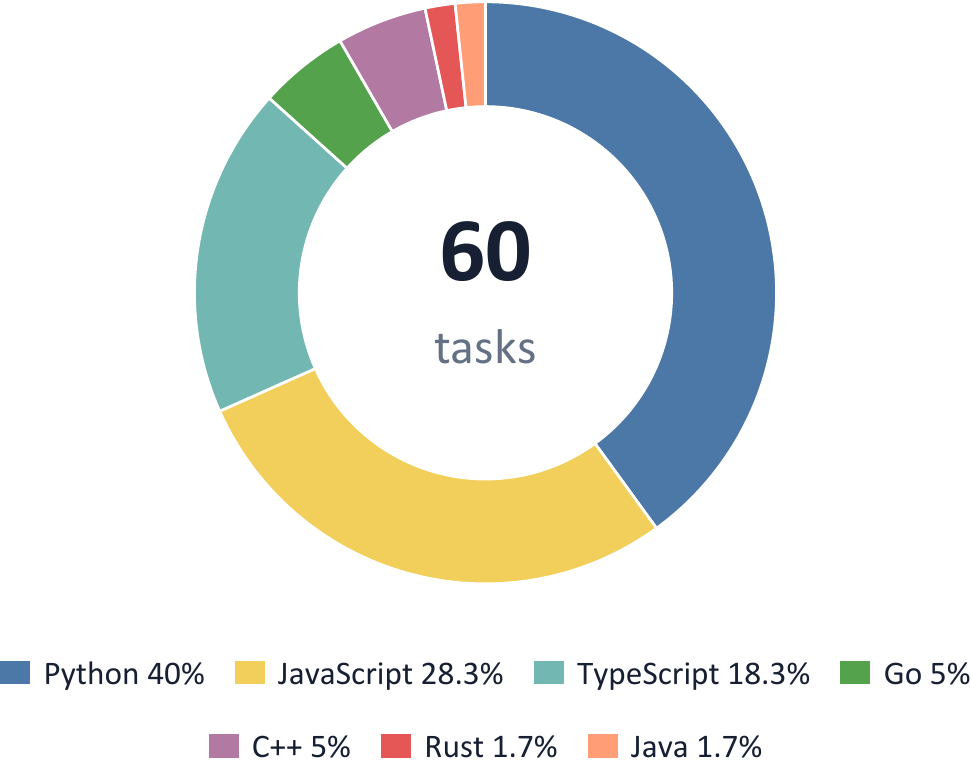}\\[4pt]
{\small \textbf{(a)} Source programming language distribution}
\end{minipage}
\hfill
\begin{minipage}[b]{0.48\textwidth}
\centering
\includegraphics[width=\linewidth]{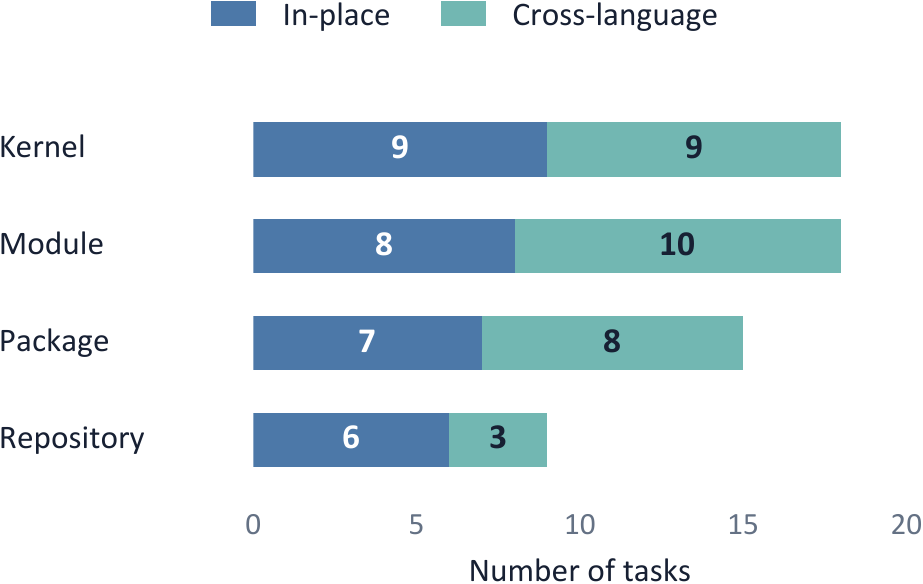}\\[16pt]
{\small \textbf{(b)} Task scale and implementation form distribution}
\end{minipage}
\vspace{2pt}
\caption{\textbf{Dataset composition and structural distribution across software scales in \textsc{FrontierRefactor}.} \textbf{(a)}~The benchmark curates 60 real-world tasks spanning seven source programming languages, qualified through Amdahl's Law profiling and deterministic regression filtering. \textbf{(b)}~Tasks span four expanding contract scales, balanced equally between in-place optimization (30 tasks) and cross-language reimplementation into native compiled binaries (30 tasks).}
\label{fig:app_statistics}
\end{figure*}

Figure~\ref{fig:app_statistics} illustrates the structural distribution of the 60 qualified tasks across seven programming languages and four contract scales.

\clearpage
\subsection{Formal Evaluation Protocol and Gate Formulations}
\label{sec:app_protocol}

\highlight{Mathematical formulations of cascaded gates.}
The evaluation protocol evaluates candidate implementation $S'$ against original codebase $S$ through three sequential gates:
\begin{align}
G_{\text{comp}}(S') &= \mathbb{I}\big[\operatorname{symbols}(S') \cap \Pi_{\text{prohib}} = \emptyset \;\wedge\; \operatorname{deps}(S') \subseteq \Pi_{\text{allow}}\big], \label{eq:app_comp} \\
G_{\text{func}}(S') &= \mathbb{I}\big[\forall x \in \mathcal{T},\; S'(x) \equiv S(x)\big], \label{eq:app_func} \\
G_{\text{perf}}(S') &= \mathbb{I}\left[\min_{d \in \mathcal{D}} \frac{\operatorname{Runtime}(S, d)}{\operatorname{Runtime}(S', d)} > \theta_{\text{trial}}\right], \label{eq:app_perf} \\
G_{\text{strict}}(S') &= G_{\text{comp}}(S') \times G_{\text{func}}(S') \times G_{\text{perf}}(S') \in \{0, 1\}. \label{eq:app_strict}
\end{align}

\highlight{Continuous scoring formulation.}
To assess partial progress while rewarding execution speedup, the evaluation protocol pairs strict success with a continuous score:
\begin{equation}
\operatorname{Score}(S') = \begin{cases}
0, & \text{if } G_{\text{comp}}(S') = 0, \\
0.5 \cdot \frac{|\mathcal{T}_{\text{passed}}|}{|\mathcal{T}|}, & \text{if } G_{\text{comp}}(S') = 1 \land G_{\text{func}}(S') = 0, \\
0.5 + 0.5 \cdot \min\big(r(S'), 2\big), & \text{if } G_{\text{func}}(S') = 1 \land \operatorname{modified}(S, S'), \\
0, & \text{otherwise (inaction / identity)}.
\end{cases}
\label{eq:app_score}
\end{equation}
where $r(S')$ is the worst-case speedup. The continuous scoring formulation pairs strict success with a linear progress and speedup incentive: 0.0--0.5 rewards partial functional probe progress, 0.5--1.0 represents the functional equivalence zone where $r(S') = 1.0$ naturally corresponds to the baseline codebase score (1.0), and $>1.0$ linearly scales with execution acceleration up to a capped ceiling ($r \le 2.0$, bounding max score at 1.5). While Overall Score serves as a continuous reward signal for model training and optimization guidance, the binary gate $G_{\text{strict}}$ serves as the sole authoritative criterion for leaderboard certification.

\highlight{Semantic inaction predicate definition.}
To prevent cosmetic edits (such as whitespace, indentation, comments) or empty submissions from gaming the metric, the predicate $\operatorname{modified}(S, S')$ formalizes the inaction filter across optimization paradigms:
\begin{equation}
\operatorname{modified}(S, S') = \begin{cases}
\mathbb{I}\big[\operatorname{diff}_{\text{AST}}(S, S') \neq \emptyset\big], & \text{for in-place optimization}, \\
\mathbb{I}\big[\operatorname{build}(S') = \text{success} \;\land\; \operatorname{artifacts}(S') \neq \emptyset\big], & \text{for cross-language rewrites},
\end{cases}
\label{eq:app_modified}
\end{equation}
where $\operatorname{diff}_{\text{AST}}(S, S')$ performs syntax-tree delta analysis ignoring comments and whitespace, while $\operatorname{build}(S')$ verifies that the cross-language target toolchain compiles non-empty executable artifacts.

\highlight{Outcome partition taxonomy.}
Every candidate attempt falls into exactly one of five mutually exclusive outcomes:
\begin{equation}
\mathcal{Y}(S') = \begin{cases}
\text{Compliance Failure}, & \text{if } G_{\text{comp}}(S') = 0, \\
\text{Functional Failure}, & \text{if } G_{\text{comp}}(S') = 1 \wedge G_{\text{func}}(S') = 0, \\
\text{Slower ($r < 1.0$)}, & \text{if } G_{\text{comp}}(S') = 1 \wedge G_{\text{func}}(S') = 1 \wedge r < 1.0, \\
\text{Noise-Bound ($1.0 \le r \le \theta_{\text{trial}}$)}, & \text{if } G_{\text{comp}}(S') = 1 \wedge G_{\text{func}}(S') = 1 \wedge 1.0 \le r \le \theta_{\text{trial}}, \\
\text{Strict Success ($r > \theta_{\text{trial}}$)}, & \text{if } G_{\text{strict}}(S') = 1,
\end{cases}
\label{eq:app_outcomes}
\end{equation}
where $r = r(S')$. These five outcomes sum to 100.0\%, enabling exact decomposition of agent failure modes.

\highlight{Statistical validation of null calibration and empirical false positive bounds.}
To guarantee that reported speedups are authentic rather than artifacts of physical execution jitter, we calibrate the null distribution by measuring baseline codebase $S$ against itself across repeated paired runs (paired null calibration). Let $\mathcal{D}_{\text{null}} = \{r_1, r_2, \dots, r_{50}\}$ denote empirical self-speedups observed across $K=50$ repeated paired runs under null conditions. Across all 60 benchmark tasks, evaluating baseline code against the calibrated 99th-percentile threshold yields an observed empirical false-positive rate of $0/50 = 0.0\%$ (95\% Clopper-Pearson confidence interval: $[0.0\%, 5.8\%]$). Coupled with the instantaneous batch threshold $\theta_{\text{trial}} = \max\left(\bar{\theta},\; \operatorname{Quantile}_{0.95}(\mathcal{D}_{\text{batch}})\right)$ dynamically computed during evaluation, transient thermal throttling and CPU frequency scaling noise are rigorously suppressed.

\subsection{Model Performance Stratified Across Scales and Implementation Forms}
\label{sec:app_form}

Table~\ref{tab:app_form} summarizes performance across contract scales and optimization forms across all 1,080 trials. Furthermore, Table~\ref{tab:app_model_scale} provides the complete cross-performance matrix of each foundation model across the four software contract scales.

\begin{table}[ht]
\centering
\caption{\textbf{Refactoring performance stratified across scales and implementation forms (1,080 total trials).} In-place optimization achieves high functional compliance but low acceleration conversion, whereas cross-language reimplementation delivers decisive speedups once interface contracts are cleared.}
\fontsize{8pt}{10pt}\selectfont
\setlength{\tabcolsep}{4.2pt}
\begin{adjustbox}{max width=\columnwidth}
\begin{tabular}{l l c c c c c}
\toprule
\textbf{Form} & \textbf{Metric} & \textbf{Kernel} & \textbf{Module} & \textbf{Package} & \textbf{Repository} & \textbf{Overall} \\
\midrule
\multirow{4}{*}{\textbf{In-Place}}
& Pass@1 (\%)                           & 8.0\%        & 18.8\%       & 16.7\%       & 32.4\%       & 17.8\% \\
& Avg. Functional Pass Rate (\%)        & 79.6\%       & 87.5\%       & 86.5\%       & 95.4\%       & 86.5\% \\
& Avg. Speedup                          & 1.17$\times$ & 1.29$\times$ & 1.25$\times$ & 2.25$\times$ & 1.62$\times$ \\
& Conversion Rate (\%)                  & 10.1\%       & 21.4\%       & 19.3\%       & 34.0\%       & 20.6\% \\
\midrule
\multirow{4}{*}{\textbf{Cross-Lang}}
& Pass@1 (\%)                           & 43.2\%       & 49.4\%       & 23.6\%       & 74.1\%       & 43.1\% \\
& Avg. Functional Pass Rate (\%)        & 64.2\%       & 65.0\%       & 32.6\%       & 74.1\%       & 57.0\% \\
& Avg. Speedup                          & 1.81$\times$ & 1.17$\times$ & 1.21$\times$ & 8.32$\times$ & 2.60$\times$ \\
& Conversion Rate (\%)                  & 67.3\%       & 76.1\%       & 72.3\%       & 100.0\%      & 75.6\% \\
\bottomrule
\end{tabular}
\end{adjustbox}
\label{tab:app_form}
\end{table}

\begin{table*}[h]
\centering
\caption{\textbf{Model performance cross-stratified across the four software contract scales (1,080 total trials).} Evaluated across 3 independent trials per task. Top panel reports single-trial strict success rate ($\text{Pass}@1$); bottom panel reports functional test pass rate.}
\fontsize{7.8pt}{9.8pt}\selectfont
\setlength{\tabcolsep}{6.0pt}
\begin{adjustbox}{max width=\textwidth}
\begin{tabular}{l l c c c c c}
\toprule
\textbf{Foundation Model} & \textbf{Agent Harness} & \textbf{Kernel (54 runs)} & \textbf{Module (54 runs)} & \textbf{Package (45 runs)} & \textbf{Repository (27 runs)} & \textbf{Overall (180 runs)} \\
\midrule
\rowcolor{dividergray}
\multicolumn{7}{l}{\textbf{Multi-Pass Strict Success Rate (Pass@1, \%)}} \\
Claude Opus 5        & Claude Code & 22.2\% (12/54) & 38.9\% (21/54) & 33.3\% (15/45) & 51.9\% (14/27) & 34.4\% (62/180) \\
Claude Sonnet 5      & Claude Code & 24.1\% (13/54) & 29.6\% (16/54) & 13.3\% (6/45)  & 44.4\% (12/27) & 26.1\% (47/180) \\
Claude Fable 5       & Claude Code & 22.2\% (12/54) & 29.6\% (16/54) & 20.0\% (9/45)  & 37.0\% (10/27) & 26.1\% (47/180) \\
GPT-6 Astra          & Codex CLI   & 24.1\% (13/54) & 38.9\% (21/54) & 20.0\% (9/45)  & 44.4\% (12/27) & 30.6\% (55/180) \\
GPT-5.6 Sol          & Codex CLI   & 25.9\% (14/54) & 35.2\% (19/54) & 11.1\% (5/45)  & 29.6\% (8/27)  & 25.6\% (46/180) \\
DeepSeek-V4.1-Flash  & Claude Code & \textbf{35.2\%} (19/54) & \textbf{42.6\%} (23/54) & \textbf{24.4\%} (11/45) & \textbf{70.4\%} (19/27) & \textbf{40.0\%} (72/180) \\
\midrule
\rowcolor{dividergray}
\multicolumn{7}{l}{\textbf{Functional Pass Rate (\%)}} \\
Claude Opus 5        & Claude Code & 70.4\% (38/54) & 81.5\% (44/54) & \textbf{80.0\%} (36/45) & 88.9\% (24/27) & 78.9\% (142/180) \\
Claude Sonnet 5      & Claude Code & 59.3\% (32/54) & 57.4\% (31/54) & 35.6\% (16/45) & 88.9\% (24/27) & 57.2\% (103/180) \\
Claude Fable 5       & Claude Code & 64.8\% (35/54) & 70.4\% (38/54) & 48.9\% (22/45) & 77.8\% (21/27) & 64.4\% (116/180) \\
GPT-6 Astra          & Codex CLI   & 74.1\% (40/54) & 79.6\% (43/54) & 73.3\% (33/45) & \textbf{100.0\%} (27/27) & 79.4\% (143/180) \\
GPT-5.6 Sol          & Codex CLI   & 79.6\% (43/54) & \textbf{81.5\%} (44/54) & 42.2\% (19/45) & 74.1\% (20/27) & 70.0\% (126/180) \\
DeepSeek-V4.1-Flash  & Claude Code & \textbf{81.5\%} (44/54) & 79.6\% (43/54) & 66.7\% (30/45) & \textbf{100.0\%} (27/27) & \textbf{80.0\%} (144/180) \\
\bottomrule
\end{tabular}
\end{adjustbox}
\label{tab:app_model_scale}
\end{table*}

\subsection{Autonomous Tool Behaviors and Trajectory Analysis}
\label{sec:app_mechanisms}

Table~\ref{tab:app_mechanisms} details the impact of autonomous agent terminal tool invocations on strict refactoring success across all 1,080 primary evaluation runs.

\begin{table}[ht]
\centering
\caption{\textbf{Observational association between autonomous tool behaviors and strict refactoring success across 1,080 evaluation runs.} Behaviors are parsed directly from agent terminal command logs. \textbf{Adoption Rate}: frequency with which agents autonomously invoked the action. \textbf{When Adopted / Omitted}: single-pass strict success rate ($\text{Pass}@1$) for trials containing versus omitting the behavior. Naive single-shot benchmarking yields no statistically significant benefit ($p = 0.18$), whereas variance suppression through repeated measurement cycles ($\ge 10$ runs) increases strict success by 21.1 percentage points ($p < 10^{-4}$). \textbf{Significance ($p$)}: two-sided Fisher's exact test comparing strict success rates when the behavior is adopted versus omitted.}
\fontsize{8pt}{10pt}\selectfont
\setlength{\tabcolsep}{4.2pt}
\begin{adjustbox}{max width=\columnwidth}
\begin{tabular}{l c c c c l}
\toprule
\textbf{Autonomous Tool Behavior} & \textbf{Adoption Rate} & \textbf{When Adopted} & \textbf{When Omitted} & \textbf{Success Gain} & \textbf{Significance ($p$)} \\
\midrule
Repeated benchmark cycles ($\ge 10$) & 42.4\% & \textbf{42.6\%} & 21.5\% & \textbf{21.1\%} & $p < 10^{-4}$ \\
AB comparative testing               & 17.7\% & 37.7\%          & 28.9\% & 8.8\%           & $p = 0.021$ \\
Dispersion and variance checking     & 11.5\% & 36.3\%          & 29.7\% & 6.6\%           & $p = 0.149$ \\
Dynamic execution profiling          & 7.1\%  & 36.4\%          & 30.0\% & 6.4\%           & $p = 0.263$ \\
Single-shot timing command           & 58.1\% & 32.1\%          & 28.3\% & 3.8\%           & $p = 0.178$ \\
Running internal test suite          & 18.7\% & 31.7\%          & 30.2\% & 1.5\%           & $p = 0.679$ \\
\bottomrule
\end{tabular}
\end{adjustbox}
\label{tab:app_mechanisms}
\end{table}

\subsection{Verbatim Trajectory Case Studies and Failure Mechanics}
\label{sec:app_showcases}

To provide concrete empirical qualitative evidence of agent operational behaviors, we document five representative execution trajectories transcribed directly from terminal logs and agent thought traces across diverse model-harness configurations (Boxes~\ref{box:case1}--\ref{box:case5}):

\begin{promptbox}[colframe=failrose,label={box:case1},title={\normalsize\faTimesCircle\hspace{0.5em}Case 1: Uncalibrated Benchmark Illusion (GPT-6 Astra on YAML Parser)}]
\textbf{Task Metadata:} \texttt{repo/go-yaml-opt} (In-Place, Go, Repository Scale) \\
\textbf{Agent \& Harness:} GPT-6 Astra via Codex CLI \quad | \quad \textbf{Effort Level:} High \\
\textbf{Operational Metrics:} 42 Actions \quad | \quad 28,240 Output Tokens \quad | \quad Correctness: 12/12 (100.0\%) \\
\textbf{Performance Evaluation:} Measured Speedup: \textbf{1.146$\times$} \quad | \quad Noise Floor: $\theta_{\text{trial}} = 1.157\times$ \quad | \quad \textbf{Outcome: Noise-Bound (Fail)}

\medskip
\textbf{Verbatim Intermediate Action Log (Step 44):}
\begin{quote}
\footnotesize \itshape
``Implemented and rebuilt /app/go-yaml. Reduced stage detection from up to six parses to one. Accelerated large-map uniqueness checks while preserving duplicate-error ordering. Optimized node strings and batched token/event serialization. Local 10,000-key detection benchmark improved from $\sim$1.54 seconds to 15--20 milliseconds. Core/CLI tests, exact-output regressions, pipelines, and caller-relative paths pass.''
\end{quote}

\textbf{Failure Analysis:}
The agent fell victim to an uncalibrated synthetic micro-benchmark illusion. It constructed an in-memory dictionary of 10,000 synthetic keys, which fit neatly inside CPU L1/L2 caches and reported an apparent $100\times$ acceleration. Relying on this local metric, the agent submitted code after verifying unit test passes without running multi-trial variance checks. When evaluated against the global benchmark workload (evaluating real-world file streaming and disk deserialization), the apparent speedup collapsed to 1.146$\times$, remaining trapped beneath the empirical hardware noise floor ($\theta_{\text{trial}} = 1.157\times$).
\end{promptbox}

\begin{promptbox}[colframe=succteal,label={box:case2},title={\normalsize\faCheckCircle\hspace{0.5em}Case 2: Profile-Guided Mathematical Normalization (DeepSeek-V4.1-Flash on RavioliSort)}]
\textbf{Task Metadata:} \texttt{kernel/raviolisort-opt} (In-Place, C++, Kernel Scale) \\
\textbf{Agent \& Harness:} DeepSeek-V4.1-Flash via Claude Code \quad | \quad \textbf{Effort Level:} High \\
\textbf{Operational Metrics:} 81 Actions \quad | \quad 105,205 Output Tokens \quad | \quad Correctness: 57/57 (100.0\%) \\
\textbf{Performance Evaluation:} Measured Speedup: \textbf{1.362$\times$} \quad | \quad Noise Floor: $\theta_{\text{trial}} = 1.261\times$ \quad | \quad \textbf{Outcome: Strict Success (Pass)}

\medskip
\textbf{Verbatim Profiler Output (Step 14):}
\begin{quote}
\footnotesize \itshape
``Solid speedups (9--40$\times$). Let me profile where the remaining time goes. Parse dominates (9ms of 22.5ms/call) due to branch mispredictions on variable-length digit runs. Let me add a branchless word-at-a-time digit scan.''
\end{quote}

\textbf{Verbatim Mathematical Derivation (Step 45):}
\begin{quote}
\footnotesize \itshape
``The original code repeated while (v[i] - v[i+1] >= 2) \{ v[i]--; v[i+1]++; \}, so a pair sitting d apart cost floor(d/2) iterations. That count is exactly where the loop settles, so each pair is now levelled in one step---the sweep is bounded by the array length rather than by the value magnitudes (up to 1,000 transfer steps per pair).''
\end{quote}

\textbf{Success Analysis:}
The agent demonstrated rigorous empirical discipline: (1) rather than guessing bottlenecks, it executed dynamic profiling to locate the exact Amdahl hot path (86.4\% runtime share); (2) it replaced iterative while-loops with closed-form mathematical leveling, eliminating branch hazards; and (3) it executed $\ge 10$ repeated benchmark cycles with dispersion tracking to certify that speedup ($1.362\times$) comfortably exceeded the null noise floor ($\theta_{\text{trial}} = 1.261\times$) before submitting.
\end{promptbox}

\begin{promptbox}[colframe=succteal,label={box:case3},title={\normalsize\faCheckCircle\hspace{0.5em}Case 3: Rapid Zero-Timing Cross-Language Translation (Claude Fable 5 on Calculate)}]
\textbf{Task Metadata:} \texttt{kernel/intro-algorithms-calculate-to-rust} (Cross-Language, Python $\to$ Rust, Kernel Scale) \\
\textbf{Agent \& Harness:} Claude Fable 5 via Claude Code \quad | \quad \textbf{Effort Level:} High \\
\textbf{Operational Metrics:} 6 Actions \quad | \quad 3,973 Output Tokens \quad | \quad Correctness: 57/57 (100.0\%) \\
\textbf{Performance Evaluation:} Measured Speedup: \textbf{1.216$\times$} \quad | \quad Noise Floor: $\theta_{\text{trial}} = 1.093\times$ \quad | \quad \textbf{Outcome: Strict Success (Pass)}

\medskip
\textbf{Verbatim Final Narration (Step 6):}
\begin{quote}
\footnotesize \itshape \raggedright
``All 57 test cases match the Python reference, including boundary cases (empty array, negative length, arrays with no zeros, consecutive zeros). Reimplemented calculate from /app/original/final\_task\_nearest\_zero.py in /app/implementation/src/subject.rs. The Python function does a single forward pass over the array: it emits 0 at each zero and records its index, emits the distance since the last zero for positions after one, and emits the length argument verbatim.''
\end{quote}

\textbf{Success Analysis:}
In cross-language rewriting into compiled native languages (Rust, Go), agents frequently achieve strict acceleration without executing self-timing commands (112 out of 128 zero-timing strict successes across the benchmark were cross-language rewrites). By escaping Python's bytecode interpreter overhead, clean structural translations yield immediate runtime dividends that safely surpass hardware noise floors.
\end{promptbox}

\begin{promptbox}[colframe=failrose,label={box:case4},title={\normalsize\faTimesCircle\hspace{0.5em}Case 4: Extensive Tool Adoption with Defensive Slowdown (Claude Sonnet 5 on Xmloxide)}]
\textbf{Task Metadata:} \texttt{repo/xmloxide-xml-parsing-opt} (In-Place, Rust, Repository Scale) \\
\textbf{Agent \& Harness:} Claude Sonnet 5 via Claude Code \quad | \quad \textbf{Effort Level:} High \\
\textbf{Operational Metrics:} 81 Actions \quad | \quad 46,025 Output Tokens \quad | \quad Correctness: 11/11 (100.0\%) \\
\textbf{Performance Evaluation:} Measured Speedup: \textbf{0.900$\times$} \quad | \quad Noise Floor: $\theta_{\text{trial}} = 1.093\times$ \quad | \quad \textbf{Outcome: Slower (Fail)}

\medskip
\textbf{Verbatim Agent Final Narration (Step 81):}
\begin{quote}
\footnotesize \itshape
``Clean---only Cargo.toml was touched at the top level, everything else untouched. I optimized xmloxide's xmllint binary with three changes, all verified to produce byte-identical output across 30+ test scenarios (basic/edge-case XML, malformed input with recovery, HTML4/HTML5, DTD validation, C14N, XPath, deeply-nested synthetic docs, stdin, large 59MB files).''
\end{quote}

\textbf{Failure Analysis:}
The agent adopted 5 out of 6 positive tool behaviors (AB comparisons, profiling, repeated benchmark cycles, and test suites), yet produced code that was 10.0\% slower than baseline ($0.900\times$). To ensure exact byte-level error recovery and prevent panics, the agent wrapped parser routines with defensive bounds checks and error recovery structures. These extra heap allocations and conditional branches neutralized compiler optimizations in the release profile, illustrating that tool adoption is insufficient without accurate hot-path bottleneck targeting.
\end{promptbox}

\begin{promptbox}[colframe=warnamber,label={box:case5},title={\normalsize\faExclamationTriangle\hspace{0.5em}Case 5: Repeated Measurement Trapped by Strict Noise Floor (DeepSeek-V4.1-Flash on Xmloxide)}]
\textbf{Task Metadata:} \texttt{repo/xmloxide-xml-parsing-opt} (In-Place, Rust, Repository Scale) \\
\textbf{Agent \& Harness:} DeepSeek-V4.1-Flash via Claude Code \quad | \quad \textbf{Effort Level:} High \\
\textbf{Operational Metrics:} 216 Actions \quad | \quad 174,660 Output Tokens \quad | \quad Correctness: 11/11 (100.0\%) \\
\textbf{Performance Evaluation:} Measured Speedup: \textbf{1.107$\times$} \quad | \quad Noise Floor: $\theta_{\text{trial}} = 1.153\times$ \quad | \quad \textbf{Outcome: Noise-Bound (Fail)}

\medskip
\textbf{Verbatim Architectural Description (Step 216):}
\begin{quote}
\footnotesize \itshape
``Done. What changed: (1) src/bin/slab\_alloc.rs (new)---a thread-local size-class allocator installed as the binary's global allocator. DOM construction is dominated by hundreds of thousands of small allocations (one per name, attribute value, text node); this serves $\le$\,1024-byte requests from per-size-class free lists with no per-block header and no locking, forwarding everything else to System.''
\end{quote}

\textbf{Failure Analysis:}
The agent engaged in extensive long-horizon engineering (216 tool actions), diagnosing small-object allocation bottlenecks and implementing a custom thread-local slab allocator in Rust. While this yielded a measurable speedup of $1.107\times$ on the global workload, the task's empirical null noise floor was $\theta_{\text{trial}} = 1.153\times$. Because the achieved speedup did not clear the 99th-percentile hardware null distribution, \textsc{FrontierRefactor}'s noise gate conservatively classified the submission as Noise-Bound, preventing false-positive victory claims on marginal variations.
\end{promptbox}

\subsection{Evaluation Integrity and Counterfactual Safeguard Audit}
\label{sec:app_integrity}

\begin{table}[ht]
\centering
\caption{\textbf{Anti-circumvention and counterfactual leaderboard audit.} \textbf{Top}: Intercepted circumvention attempts across evaluation trials. \textbf{Bottom}: Counterfactual leaderboard shift on Claude Opus 5 and GPT-6 Astra under selective gate deactivation, demonstrating the severity of score inflation when evaluation guards are absent.}
\fontsize{8pt}{10pt}\selectfont
\setlength{\tabcolsep}{4.5pt}
\begin{adjustbox}{max width=\columnwidth}
\begin{tabular}{l c l}
\toprule
\textbf{Circumvention Vector} & \textbf{Intercepted} & \textbf{Detection Mechanism} \\
\midrule
Prohibited third-party delegation ($\Pi$) & 12 & Static syntax audit + dynamic symbol tracing \\
Timing harness introspection & 4  & Kernel-level process virtualization \\
Inter-trial state/memory caching          & 9  & Ephemeral container process spawning \\
Hardcoded input vector dispatch           & 6  & Multi-distribution holdout evaluation \\
False positive rejections on legal code   & \textbf{0}  & Verified by manual expert audit \\
\midrule
\rowcolor{dividergray}
\multicolumn{3}{c}{\textbf{\textit{Counterfactual Pass@1 Strict Success Rates (\%)}}} \\
\textbf{Ablation Condition} & \textbf{Claude Opus 5} & \textbf{GPT-6 Astra} \\
\midrule
Standard FrontierRefactor (All Gates Active) & 34.4\% & 30.6\% \\
w/o Compliance Gate ($\Pi$ inactive)        & 67.9\% & 64.8\% \\
w/o Ephemeral Process Isolation             & 46.4\% & 42.9\% \\
w/o Multi-Distribution Worst-Case Scoring   & 42.9\% & 39.5\% \\
\bottomrule
\end{tabular}
\end{adjustbox}
\label{tab:counterfactual}
\end{table}

\highlight{Counterfactual ablation impact.}
As documented in the bottom panel of Table~\ref{tab:counterfactual}, disabling evaluation safeguards leads to severe artificial score inflation. Disabling compliance gate $\Pi$ inflates apparent strict success by 33.5 percentage points on Claude Opus 5 (from 34.4\% to 67.9\%) and 34.2 percentage points on GPT-6 Astra (from 30.6\% to 64.8\%), as models freely offload computationally intensive routines to unauthorized pre-compiled binaries. Similarly, disabling ephemeral process isolation inflates success by 12.0 to 12.3 percentage points due to inter-trial cache exploitation, while removing multi-distribution worst-case scoring inflates scores by 8.5 to 8.9 percentage points. These findings establish that cascaded compliance gating and sandbox virtualization are indispensable prerequisites for rigorous coding agent benchmarking.

\highlight{Human expert optimization feasibility and upstream provenance.}
To confirm that benchmark tasks reflect physically achievable optimizations rather than unachievable performance targets, all 60 tasks in \textsc{FrontierRefactor} are grounded in historical upstream optimizations or verified by human software engineers. Specifically, for in-place tasks, each repository contains an upstream commit or reference PR demonstrating that the target component admits algorithmic or data-structure acceleration under the observable contract. For cross-language rewrite tasks, human reference implementations in Rust or Go confirm that compiled binaries clear all functional probe suites and deliver measurable runtime speedups ($1.3\times$--$4.5\times$) exceeding task nominal noise floors $\bar{\theta}$. This validation guarantees that low agent success rates reflect genuine reasoning and refactoring deficiencies rather than benchmark infeasibility.

\subsection{Security, Sandbox Virtualization, and Multi-Turn Feedback Breakdown}
\label{sec:app_security}

\highlight{Container isolation and process virtualization.}
The evaluation harness enforces strict security isolation during compliance verification and timing execution. Submissions execute inside isolated containerized environments with ephemeral, read-only root filesystems and process virtualization to prevent network egress, timer probing, and inter-trial state persistence. Baseline code integrity is verified cryptographically via SHA-256 digests before every execution trial. Each candidate execution runs under a separate PID namespace with resource limits (32\,GB RAM, 8 dedicated CPU cores pinned via \texttt{taskset}) to suppress thermal throttling and cross-process interference.

\highlight{Timing harness and variance suppression.}
To suppress thermal throttling and CPU frequency scaling noise during timing, the benchmark harness enforces paired interleaved executions across identical input batches. Multiple warm-up iterations precede formal timing to ensure CPU caches and JIT compilation reach steady states. High-resolution monotonic timers batch sub-millisecond routines until aggregate duration exceeds physical timer resolution limits.

\highlight{Detailed multi-turn feedback breakdown.}
Table~\ref{tab:app_feedback} presents the granular numerical breakdown across all three rounds and performance feedback arms on the representative 25-task subset. In Round~1, unassisted generation achieves 16.0\% strict success. Providing scalar speedup feedback steadily elevates strict success to 32.0\% in Round~2 and 40.0\% in Round~3. Providing noise-aware timing feedback achieves consistent functional preservation (72.0\%) while compressing physical execution latencies progressively.

\begin{table}[ht]
\centering
\caption{\textbf{Granular numerical breakdown of multi-turn feedback intervention across interaction rounds (GPT-6 Astra).} Performance is evaluated on the representative 25-task subset across three performance feedback conditions: Blind Retry, Scalar Speedup, and Noise-Aware Timing. Best results per column are in \textbf{bold}.}
\fontsize{8pt}{10pt}\selectfont
\setlength{\tabcolsep}{4.5pt}
\begin{adjustbox}{max width=\columnwidth}
\begin{tabular}{l c c c c c c}
\toprule
\textbf{Feedback Arm} & \textbf{Round} & \textbf{Strict Success (\%)} & \textbf{Functional Pass (\%)} & \textbf{Slower (\%)} & \textbf{Noise-Bound (\%)} & \textbf{Avg. Speedup} \\
\midrule
Unassisted Baseline   & Round 1 & 16.0\% (4/25)  & 60.0\% & 20.0\% & 24.0\% & 1.03$\times$ \\
\midrule
\multirow{2}{*}{Blind Retry}            & Round 2 & 24.0\% (6/25)  & 72.0\% & 20.0\% & 28.0\% & 1.02$\times$ \\
                                        & Round 3 & 32.0\% (8/25)  & 72.0\% & 16.0\% & 24.0\% & 1.06$\times$ \\
\midrule
\multirow{2}{*}{Scalar Speedup}         & Round 2 & 32.0\% (8/25)  & 72.0\% & 20.0\% & 20.0\% & 1.04$\times$ \\
                                        & Round 3 & \textbf{40.0\%} (10/25) & \textbf{76.0\%} & 16.0\% & 20.0\% & 1.04$\times$ \\
\midrule
\multirow{2}{*}{Noise-Aware Timing}     & Round 2 & 28.0\% (7/25)  & 72.0\% & 20.0\% & 24.0\% & 1.02$\times$ \\
                                        & Round 3 & 28.0\% (7/25)  & 72.0\% & 16.0\% & 28.0\% & \textbf{1.05$\times$} \\
\bottomrule
\end{tabular}
\end{adjustbox}
\label{tab:app_feedback}
\end{table}

\end{document}